\documentclass{article}

\usepackage{microtype}
\usepackage{graphicx}
\usepackage{subfigure}
\usepackage{booktabs} 

\usepackage{hyperref}

\usepackage[accepted]{icml2020}

\usepackage{ulem}
\usepackage{mathtools}
\usepackage{enumitem}
\usepackage{amsmath}
\usepackage{amssymb}
\usepackage{graphics}
\usepackage{textcomp}
\usepackage{multicol}

\newcommand{\G}{\mathcal{G}}
\newcommand{\F}{\mathcal{F}}

\newcommand{\D}{\mathcal{D}}

\newcommand{\Pd}{P_{\operatorname{data}}}
\newcommand{\Pg}{P_{\mathcal{G}}}
\newcommand{\Expect}{\mathbb{E}}
\newcommand{\Reals}{\mathbb{R}}
\newcommand{\Id}{\operatorname{Id}}

\icmltitlerunning{Implicit competitive regularization in GANs}

\begin{document}

\twocolumn[
\icmltitle{Implicit competitive regularization in GANs}



\icmlsetsymbol{equal}{*}

\begin{icmlauthorlist}
\icmlauthor{Florian Sch{\"a}fer}{equal,ca}
\icmlauthor{Hongkai Zheng}{equal,sh,vca}
\icmlauthor{Anima Anandkumar}{ca}
\end{icmlauthorlist}

\icmlaffiliation{ca}{Caltech}
\icmlaffiliation{sh}{Shanghai Jiao Tong university}
\icmlaffiliation{vca}{This work was produced while HZ was a visiting undergraduate researcher at Caltech}

\icmlcorrespondingauthor{Florian Schaefer}{schaefer@caltech.edu}
\icmlcorrespondingauthor{Hongkai Zheng}{devzhk@sjtu.edu.cn}

\icmlkeywords{Machine Learning, ICML}

\vskip 0.3in
]



\printAffiliationsAndNotice{\icmlEqualContribution} 

\begin{abstract}
The success of GANs is usually attributed to properties of the divergence obtained by an optimal discriminator.
In this work we show that this approach has a fundamental flaw:\\
If we do not impose regularity of the discriminator, it can exploit visually imperceptible errors of the generator to always achieve the maximal generator loss.
In practice, gradient penalties are used to regularize the discriminator.
However, this needs a metric on the space of images that captures visual similarity.
Such a metric is not known, which explains the limited success of gradient penalties in stabilizing GANs.\\
Instead, we argue that the implicit competitive regularization (ICR) arising from the simultaneous optimization of generator and discriminator enables GANs performance.
We show that opponent-aware modelling of generator and discriminator, as present in competitive gradient descent (CGD), can significantly strengthen ICR and thus stabilize GAN training without explicit regularization.
In our experiments, we use an existing implementation of WGAN-GP and show that by training it with CGD without any explicit regularization, we can improve the inception score (IS) on CIFAR10, without any hyperparameter tuning.
\end{abstract}


\section{Introduction}
\textbf{Generative adversarial networks (GANs):}~\citep{goodfellow2014generative} are a class of generative models based on a competitive game between a \emph{generator} that tries to generate realistic new data, and a \emph{discriminator} that tries to distinguish generated from real data. In practice, both players are parameterized by neural networks that are trained simultaneously by a variant of stochastic gradient descent. 

\textbf{The minimax interpretation:}
Presently, the success of GANs is mostly attributed to properties of the divergence or metric obtained under an optimal discriminator.
For instance, an optimal discriminator in the original GAN leads to a generator loss equal to the Jensen-Shannon divergence between real and generated distribution.
Optimization over the generator is then seen as approximately minimizing this divergence. 
We refer to this point of view as the \emph{minimax interpretation}.
The minimax interpretation has led to the development of numerous GAN variants that aim to use divergences or metrics with better theoretical properties.

\begin{figure*}
    \centering
    \includegraphics[width=0.30\textwidth]{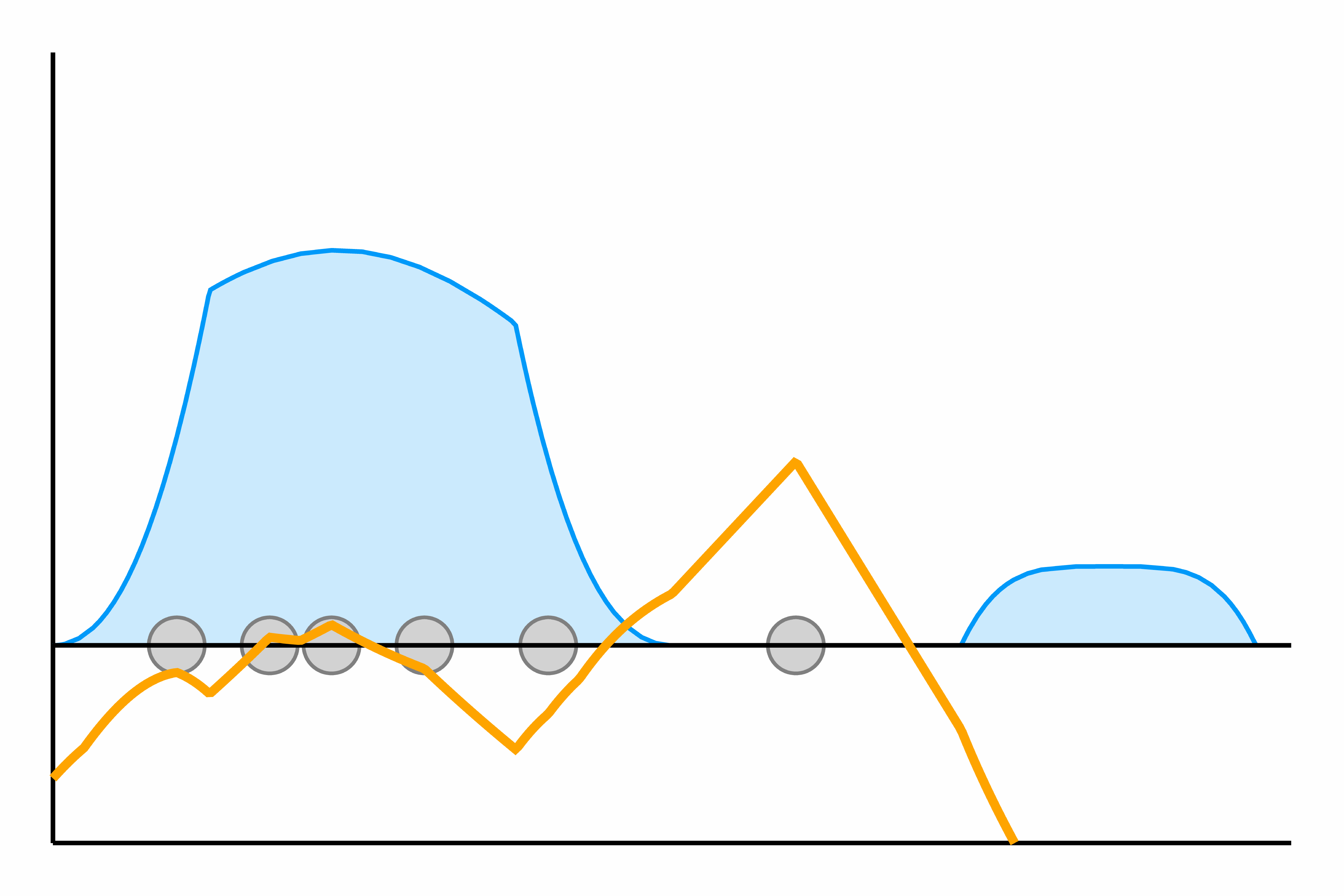}
    \hspace{0.2\textwidth}
    \includegraphics[width=0.30\textwidth]{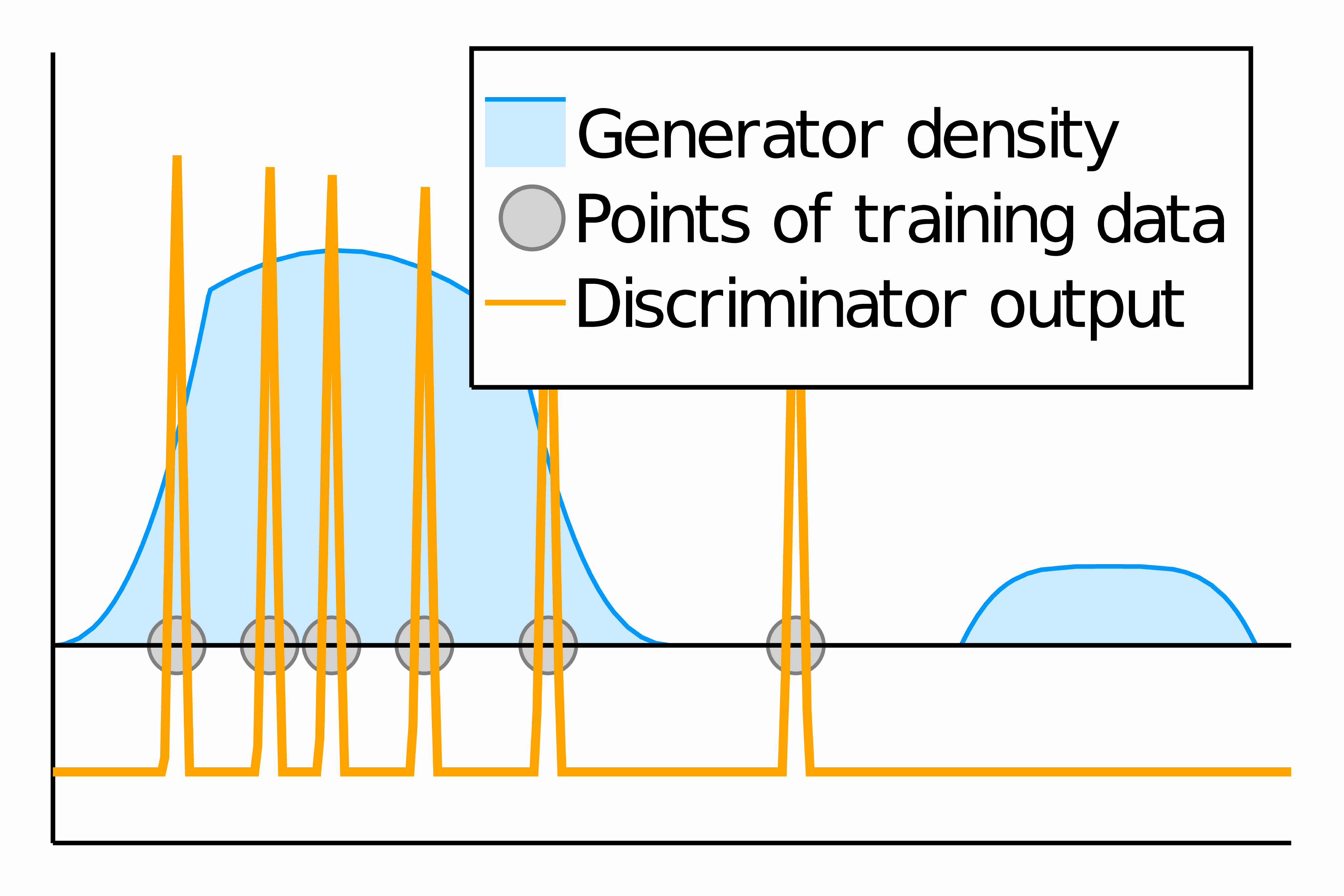}
    \caption{\textbf{The discriminator can always improve:} We want the discriminator confidence to reflect the relative abundance of true and fake data (left). But by picking out individual data points, the discriminator can almost always achieve arbitrarily low loss on any finite data set (right). 
    Even in the limit of infinite data, the slightest misalignment of the supports of generated and real data can be exploited in a similar way.}
    \label{fig:picking_out}
\end{figure*}

\textbf{The GAN-dilemma:}
However, every attempt to explain GAN performance with the minimax interpretation faces one of the two following problems:
\begin{enumerate}[wide, labelwidth=!, labelindent=0pt, label=\textbf{\arabic*.}]
\item \label{item:dilemma1} \textbf{Without regularity constraints, the discriminator can always be perfect.} This is because it can selectively assign a high score to the finite amount of real data points while assigning a low score on the remaining support of the generator distribution, as illustrated in Figure~\ref{fig:picking_out}.
Therefore, the Jensen-Shannon divergence between a continuous and a discrete distribution always achieves its maximal value, a property that is shared by all divergences that do not impose regularity constraints on the discriminator.
Thus, these divergences can not meaningfully compare the quality of different generators.
\item \label{item:dilemma2} \textbf{Imposing regularity constraints needs a measure of similarity of images.} Imposing regularity on the discriminator amounts to forcing it to map similar images to similar results. To do so, we would require a notion of similarity between images that is congruent with human perception. 
This is a longstanding unsolved problem in computer vision. 
Commonly used gradient penalties use the Euclidean norm which is known to poorly capture visual similarity, as illustrated in Figure~\ref{fig:deception}. 
\end{enumerate}
We believe that the different divergences underlying the various GAN formulations have little to do with their ability to produce realistic images.
This is supported by the large scale studies of \citet{lucic2017gans} that did not find systematic differences in the performance of GANs associated with different divergence measures.
Understanding of GAN performance is crucial in order to improve training stability and reduce the required amount of hyperparameter tuning.

\textbf{A way out?:}
Due to the GAN-dilemma, every attempt at explaining the performance of GANs needs to go beyond the minimax interpretation and consider the \emph{dynamics} of the training process.
In this work, we argue that an implicit regularization due to the simultaneous \footnote{Here and in the following, when talking about simultaneous training, we include variants such as alternating gradient descent.} training of generator and discriminator allows GANs to use the inductive biases of neural networks for the generation of realistic images.

\textbf{Implicit competitive regularization:} 
We define  \emph{implicit competitive regularization} (ICR) as the introduction of additional stable points or regions due to the simultaneous training of generator and discriminator that do not exist when only training the generator (or discriminator) with gradient descent while keeping the discriminator (or generator) fixed. \\
It has been previously observed  that performing simultaneous gradient descent (SimGD) on both players leads to stable points that are not present when performing gradient descent with respect to either player, while keeping the other player fixed \citep{mazumdar2018convergence}.
These stable points are not local Nash equilibria, meaning that they are not locally optimal for both players.
This phenomenon is commonly seen as a shortcoming of SimGD and modifications that promote convergence only to local Nash equilibria which have been proposed by, for instance, \citep{balduzzi2018mechanics,mazumdar2019finding}.
In contrast to this view we believe that ICR is crucial to overcoming the GAN-dilemma and hence to explaining GAN performance in practice by allowing the inductive biases of the discriminator network to inform the generative model.

\subsection*{Summary of Contributions}
In this work, we point out that a fundamental dilemma prevents the common minimax interpretation of GANs from explaining their successes.
We then show that implicit competitive regularization (ICR), which so far was believed to be a \emph{flaw} of SimGD, is key to overcoming this dilemma. 
Based on simple examples and numerical experiments on real GANs we illustrate how it allows to use the inductive biases of neural networks for generative modelling, resulting in the spectacular performance of GANs.\\
We then use this understanding to improve GAN performance in practice.
Interpreting ICR from a game-theoretic perspective, we reason that strategic behavior and opponent-awareness of generator and discriminator during the training procedure can strengthen ICR.
These elements are present in competitive gradient descent (CGD) \citep{schaefer2019competitive} which is based on the two players solving for a local Nash-equilibrium at each step of training. 
Accordingly, we observe that CGD greatly strengthens the effects of ICR.
In comprehensive experiments on CIFAR 10, competitive gradient descent stabilizes previously unstable GAN formulations and achieves higher inception score compared to a wide range of explicit regularizers, using both WGAN loss and the original saturating GAN loss of \citet{goodfellow2014generative}.
In particular, taking an existing WGAN-GP implementation, dropping the gradient penalty, and training with CGD leads to the highest inception score in our experiments.
We interpret this as additional evidence that ICR, as opposed to explicit regularization, is the key mechanism behind GAN performance.

\section{The GAN-dilemma}
\label{sec:gan_dilemma}

In this section, we study in more detail the fundamental dilemma that prevents the common minimax interpretation from explaining the successes of GANs.
In particular, we show how the existing GAN variants fall into one or the other side of the GAN-dilemma.

\textbf{Metric-agnostic GANs:}
In the original formulation due to \citet{goodfellow2014generative}, the two players are playing a zero-sum game with the loss function of the generator given by the binary cross entropy
\begin{figure*}
    \centering
    \includegraphics[width=0.80\textwidth]{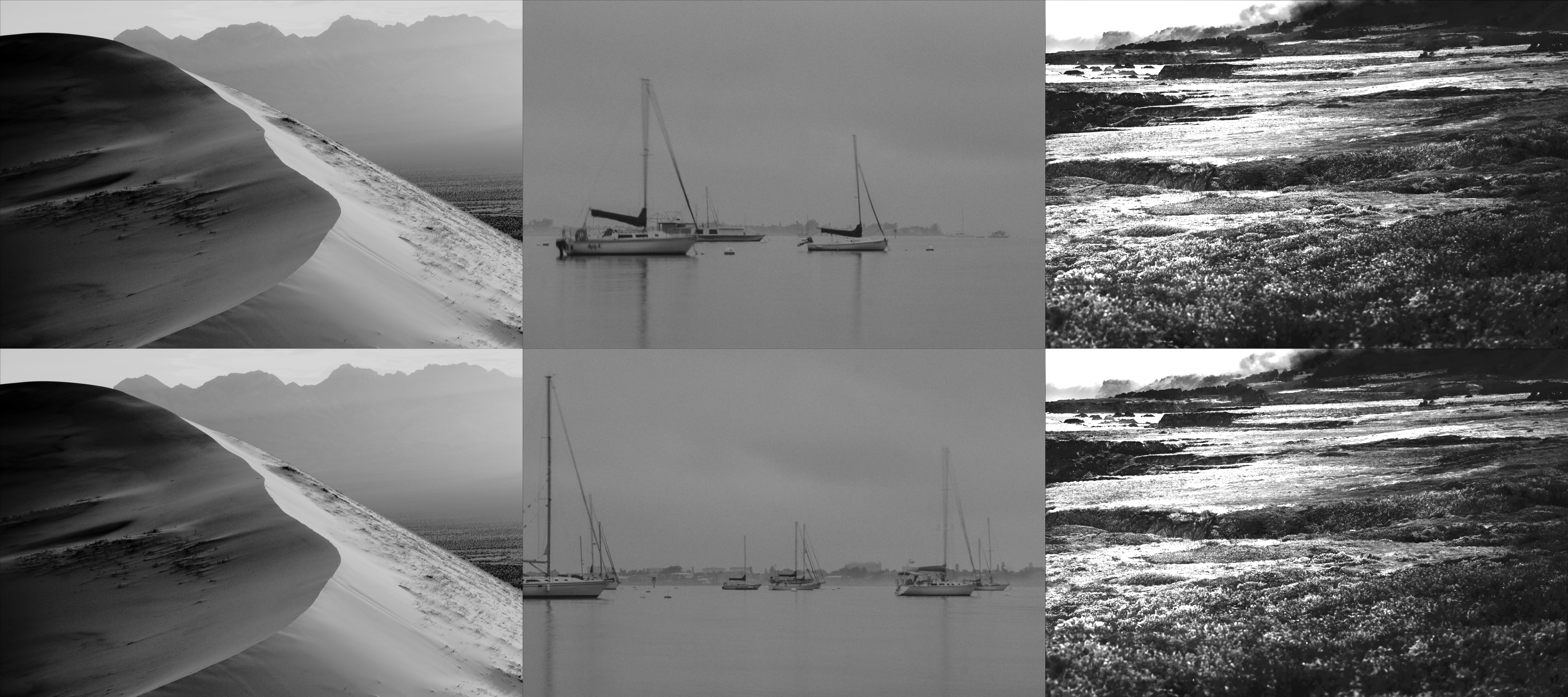}
    \caption[]{\textbf{The Euclidean distance is not perceptual:} We would like to challenge the reader to order the above three pairs of images according to the Euclidean distance of their representation as vectors of pixel-intensities. \footnotemark}
    \label{fig:deception}
\end{figure*}
\footnotetext{The pairs of images are ordered from left to right, in increasing order of distance. The first pair is identical, while the third pair differs by a tiny warping.}

\begin{equation}
    \label{eqn:ogan}
    \min \limits_{\mathcal{G}} \max \limits_{\mathcal{D}} \frac{1}{2} \Expect_{x \sim \Pd} \left[ \log \mathcal{D}(x) \right] + \frac{1}{2} \Expect_{x \sim \Pg}\left[ \log\left(1 - \D(x) \right) \right].
\end{equation}
Here, $\G$ is the probability distribution generated by the generator, $\D$ is the classifier provided by the discriminator, and $\Pd$ is the target measure, for example the empirical distribution of the training data.
A key feature of the original GAN is that it depends on the discriminator only through its output when evaluated on samples. This property is shared, for instance, by the more general class of $f$-divergence GANs \citep{nowozin2016f}.
We call GAN formulations with this property \emph{metric-agnostic}.

\textbf{Metric-informed GANs:}
To address instabilities observed in the original GAN, \citet{arjovsky2017wasserstein} introduced WGAN, with loss function given by 
\begin{equation}
    \label{eqn:wgan}
    \min \limits_{\mathcal{G}} \max \limits_{\mathcal{D}} \Expect_{x \sim \Pd} \left[ \mathcal{D}(x) \right] - \Expect_{x \sim \Pg}\left[ \D(x) \right] + \mathcal{F}\left(\nabla \mathcal{D}\right)
\end{equation}
where $\F(\nabla \D)$ is infinity if $\sup_x \left\|\nabla \D(x) \right\| > 1$ and zero, else. 
\citep{gulrajani2017improved} propose WGAN-GP, where this inequality constraint is relaxed by replacing $\mathcal{F}$ with a penalty, for instance $\F(\nabla \D) = \Expect \left[ \left( \left\|\nabla_x \D \right\| - 1 \right)^2\right]$.
These GAN formulations are fundamentally different from metric-agnostic GANs in that they depend explicitly on the gradient of the discriminator.
In particular, they depend on the choice of metric used to measure the size of $\nabla \D$.
Subsequent to WGAN(-GP), which uses the Euclidean norm, other variants such as Sobolev-GAN \citep{mroueh2017sobolev}, Banach-GAN \citep{adler2018banach}, or Besov-GAN \citep{uppal2019nonparametric} have been proposed that use different metrics to measure gradient size.
We refer to these types of GAN formulations as \emph{metric-informed} GANs.

\textbf{The problem with metric-agnostic GANs:}
GANs are able to generate highly realistic images, but they suffer from unstable training and mode collapse that often necessitates extensive hyperparameter tuning.
Beginning with \citep{arjovsky2017towards} these problems of the original GAN have been explained with the fact that the supports of the generator distribution and the training data are almost never perfectly aligned.
For any fixed generator, the discriminator can take advantage of this fact to achieve arbitrarily low loss, as illustrated in Figure~\ref{fig:picking_out}.
In the case of the Formulation \ref{eqn:ogan}, this corresponds to the well known fact that the Jensen-Shannon divergence between mutually singular measures is always maximal.
This result extends to \emph{all} metric-agnostic divergences, simply because they have no way of accessing the degree of similarity between data points on disjoint supports.

\citet{arora2017generalization,huang2017parametric} emphasize that the discriminator is restricted to a function class parameterized by a neural network. 
However, the experiments of \citet{arjovsky2017towards} as well as our own in Figure~\ref{fig:overtraining_1} clearly show the tendency of the discriminator to diverge as it achieves near-perfect accuracy.
This is not surprising since \citet{zhang2016understanding} observed that modern neural networks are able to fit even \emph{random} data perfectly.
\citet{arjovsky2017towards} also show that as the discriminator improves its classification loss, the generator achieves less and less useful gradient information.
This is again not surprising, since confidence scores of deep neural networks are known to be poorly calibrated \citep{guo2017calibration}. 
Therefore, the outputs of a near-perfect discriminator can not be expected to provide a useful assessment of the quality of the generated samples.

Since GAN optimization is highly non-convex it is natural to ask if GANs find \emph{locally} optimal points in the form of local Nash or Stackelberg equilibria.
This local minmax interpretation has been emphasized by \citet{fiez2019convergence,jin2019minmax}, but the experiments of \citet{berard2019closer} as well as our own in Figure~\ref{fig:overtraining_1} suggest that good GAN solutions for metric-agnostic GANs are typically not locally optimal for both players.
It seems plausible that the discriminator, being highly overparameterized, can find a direction of improvement against most generators.

\textbf{The problem with metric-informed GANs:}
The above observation has motivated the introduction of metric-informed GANs that restrict the size of the gradient of the discriminator (as a function mapping images to real numbers).
This limits the discriminator's ability to capitalize on small misalignments between $\D$ and $\Pd$ and thus makes for a meaningful minimax interpretation even if the two measures have fully disjoint support.
However, the Achilles heel of this approach is that it needs to choose a metric to quantify the magnitude of the discriminator's gradients.
Most of the early work on metric-informed GANs chose to measure the size of $\nabla \D$ using the Euclidean norm \citep{arjovsky2017towards,arjovsky2017wasserstein,gulrajani2017improved,roth2017stabilizing,kodali2017convergence,miyato2018spectral}.
However, since the discriminator maps images to real numbers, this corresponds to quantifying the similarity of images at least locally by the Euclidean distance of vectors containing the intensity values of each pixel.
As illustrated in Figure~\ref{fig:deception}, this notion of similarity is poorly aligned with visual similarity even locally.
From this point of view it is not surprising that the generative model of \cite{chen2019gradual}, based on a differentiable optimal transport solver, produced samples of lower visual quality than WGAN-GP, despite achieving better approximation in Wasserstein metric.
As noted by \citet{chen2019gradual}, these observations suggest that the performance of WGAN can not be explained by its relationship to the Wasserstein distance.
When comparing a variety of GAN formulations with a fixed budget for hyperparameter tuning, \citet{lucic2017gans} did not find systematic differences in their performance.
This provides additional evidence that the key to GAN performance does not lie in the choice of a particular divergence between probability measures.

The metric-informed divergences considered so far were all based on the Euclidean distance between images.
Other researchers have tried using different metrics on image space such as Sobolev or Besov norms \citep{adler2018banach,mroueh2017sobolev,uppal2019nonparametric}, or kernel maximum mean discrepancy distances \citep{li2015generative,li2017mmd,binkowski2018demystifying}.
However, none of these metrics do a good job at capturing perceptual similarity either, which explains why these variants have not been observed to outperform WGAN(-GP) in general.
Researchers in computer vision have proposed more sophisticated domain-specific distance measures \citep{simard1998transformation}, kernel functions \citep{haasdonk2007invariant,song2014local}, and features maps \citep{dalal2005histograms}.
Although computationally expensive, methods from differential geometry have been used for image inter-- and extrapolation \citep{trouve2005metamorphoses,berkels2015time,effland2018image}.
However, none of these classical methods achieve performance comparable to that of neural network based models, making them unlikely solutions for the GAN dilemma. 

\textbf{A way out:}
Generative modelling means producing new samples that are \emph{similar} to the training samples, but \emph{not too similar} to each other.
Thus, every generative method needs to choose how to measure similarity between samples, implicitly or explicitly.\\
When analyzing GANs from the minimax perspective this assessment of image similarity seems to rely exclusively on the classical metrics and divergences used for their formulation.
But modeling perceptual similarity is hard and most commonly used GAN formulations are based on measures of similarity that are known to be terrible at this task.
Thus, the minimax point of view can not explain why GANs produce images of higher visual quality than any other method.
The key to image classification is to map \emph{similar} images to \emph{similar} labels.
The fact that deep neural networks drastically outperform classical methods in this tasks leads us to believe that they capture perceptual similarity between images far better than any classical model.
We believe that the success of GANs is due to their ability to implicitly use the inductive biases of the discriminator network as a notion of similarity. 
They create images that \emph{look real} to a neural network, which acts as a proxy for \emph{looking real} to the human eye.
In the next section we propose a new mechanism, implicit competitive regularization, to explain this behavior.

\section{Implicit competitive regularization (ICR)}
\textbf{Implicit regularization:}
Based on the discussion in the last section, any attempt at understanding GANs needs to involve the inductive biases of the discriminator.
However, there is ample evidence that the inductive biases of neural networks do not arise from a limited ability to represent certain functions. 
Indeed, it is known that modern neural networks can fit almost arbitrary functions \citep{kolmogorov1956representation,cybenko1989approximation,zhang2016understanding}.
Rather, they seem to arise from the dynamics of gradient-based training that tends to converge to classifiers that generalize well, a phenomenon commonly referred to as \emph{implicit regularization} \citep{neyshabur2017implicit,gunasekar2017implicit,ma2017implicit,azizan2019stochastic,kubo2019implicit,arora2019implicit}.

\textbf{Implicit regularization is not enough for GANs:}
The implicit regularization induced by gradient descent lets neural networks prefer sets of weights with good generalization performance.
However, the outputs of even a well-trained neural network are typically not informative about the confidence of the predicted class \citep{guo2017calibration}.
Thus, a discriminator trained on finite amounts of real data and data generated by a given generator can be expected to distinguish new real data from new data generated by a similar generator, with high accuracy.
However, its outputs do not quantify the confidence of its prediction and thus of the visual quality of the generated samples.
Therefore, even considering implicit regularization, a fully trained discriminator does not provide useful gradients for training the generator.

\textbf{Implicit \emph{competitive} regularization:}
We think that GAN training relies on \emph{implicit competitive regularization} (ICR), an additional implicit regularization due to the simultaneous training of generator and discriminator.
When training generator and discriminator simultaneously, ICR selectively stabilizes good generators that would not be stable when training one player while keeping the other player fixed.

Consider the game given by 
\begin{equation}
    \label{eqn:basicICR}
    \min \limits_x \max \limits_{y} x^2 + 10 xy + y^2.
\end{equation}
In this problem, for any fixed $x$, any choice of $y$ will be sub-optimal and gradient ascent on $y$ (with $x$ fixed) will diverge to infinity for almost all initial values.

\begin{figure}
    \centering
    \includegraphics[width=0.9\columnwidth]{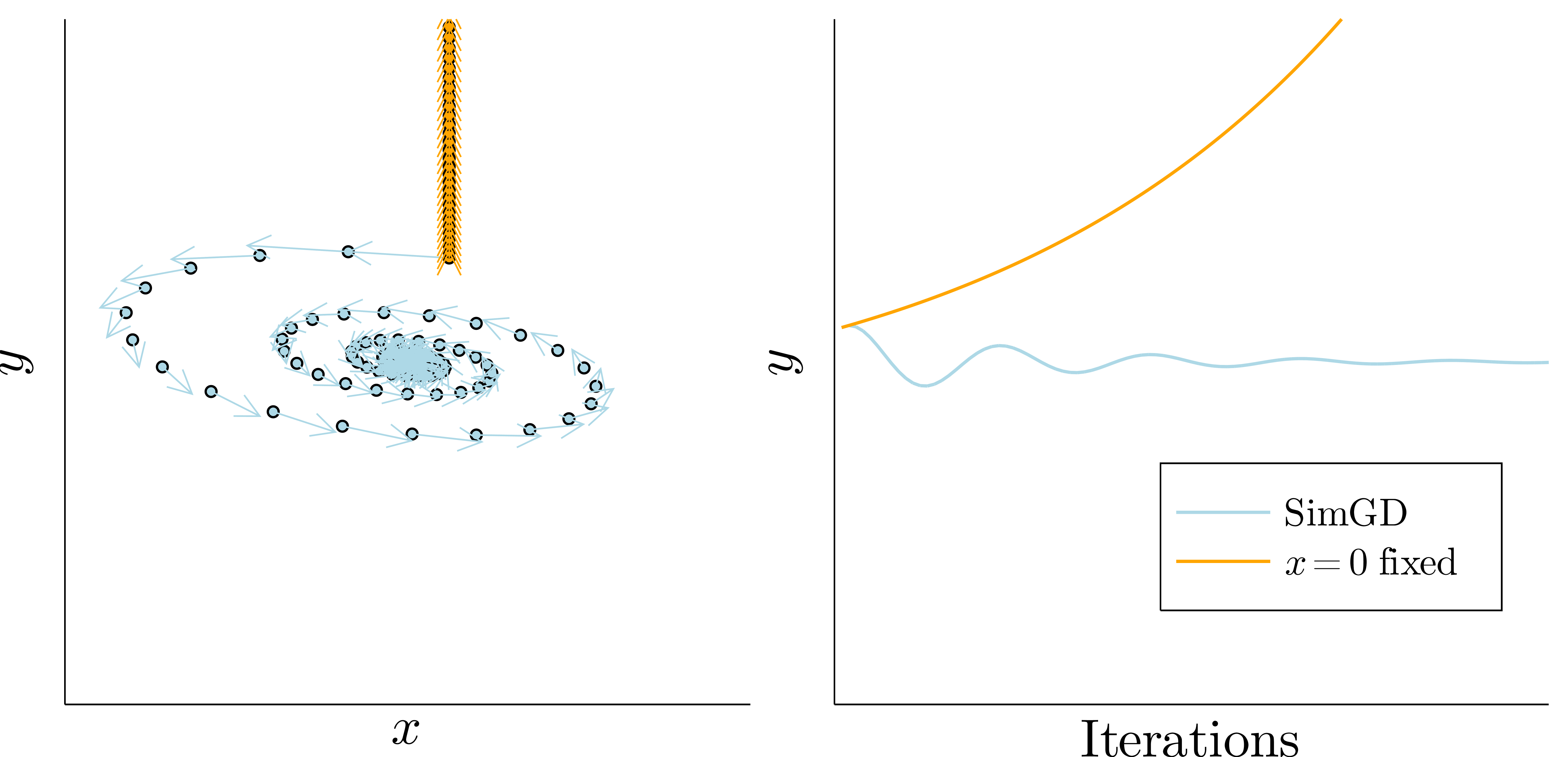}
    \caption{\textbf{ICR in the quadratic case:} When optimizing only $y$ in Equation~\eqref{eqn:basicICR}, it diverges rapidly to infinity, for any fixed $y$. If however we simultaneously optimize $x$ and $y$ with respective step sizes $\eta_{x} = 0.09$ and $\eta_{y} = 0.01$, we converge to $(0,0)$.}
    \label{fig:cycling}
\end{figure}

What about simultaneous gradient descent? 
As has been observed before \citep{mazumdar2018convergence}, simultaneous gradient descent wit step sizes $\eta_x = 0.09$ for $x$ and $\eta_y = 0.01$ for $y$ will converge to $(0,0)$, despite it being a locally \emph{worst} strategy for the maximizing player. (See Figure~\ref{fig:cycling} for an illustration.)
This is a first example of ICR, whereby the simultaneous optimization of the two agents introduces additional attractive points to the dynamics that are \emph{not} attractive when optimizing one of the players using gradient descent while keeping the other player fixed.

As outlined in Section~\ref{sec:gan_dilemma}, the key to the performance of GANs has to lie in the simultaneous optimization process.
We  now provide evidence that the solutions found by GANs are indeed stabilized by ICR.
To this end, we train a GAN on MNIST until it creates good images. 
We refer to the resulting generator and discriminator as the \emph{checkpoint} generator and discriminator.
We observe that the loss of both generator and discriminator, as well as the image quality, is somewhat stable even though it would diverge after a long time of training. 
If instead, starting at the checkpoint, we optimize only the discriminator while keeping the generator fixed, we observe that the discriminator loss drops rapidly.
For the same number of iterations and using the same learning rate, the discriminator moves away from the checkpoint significantly faster as measured both by the Euclidean norm of the weights and the output on real-- and fake images.
The observation that the discriminators diverges from the checkpoint faster when trained individually than when trained simultaneously with the generator suggests that the checkpoint, which produced good images, was stabilized by ICR.

\begin{figure}
    \centering
    \includegraphics[width=0.49\columnwidth]{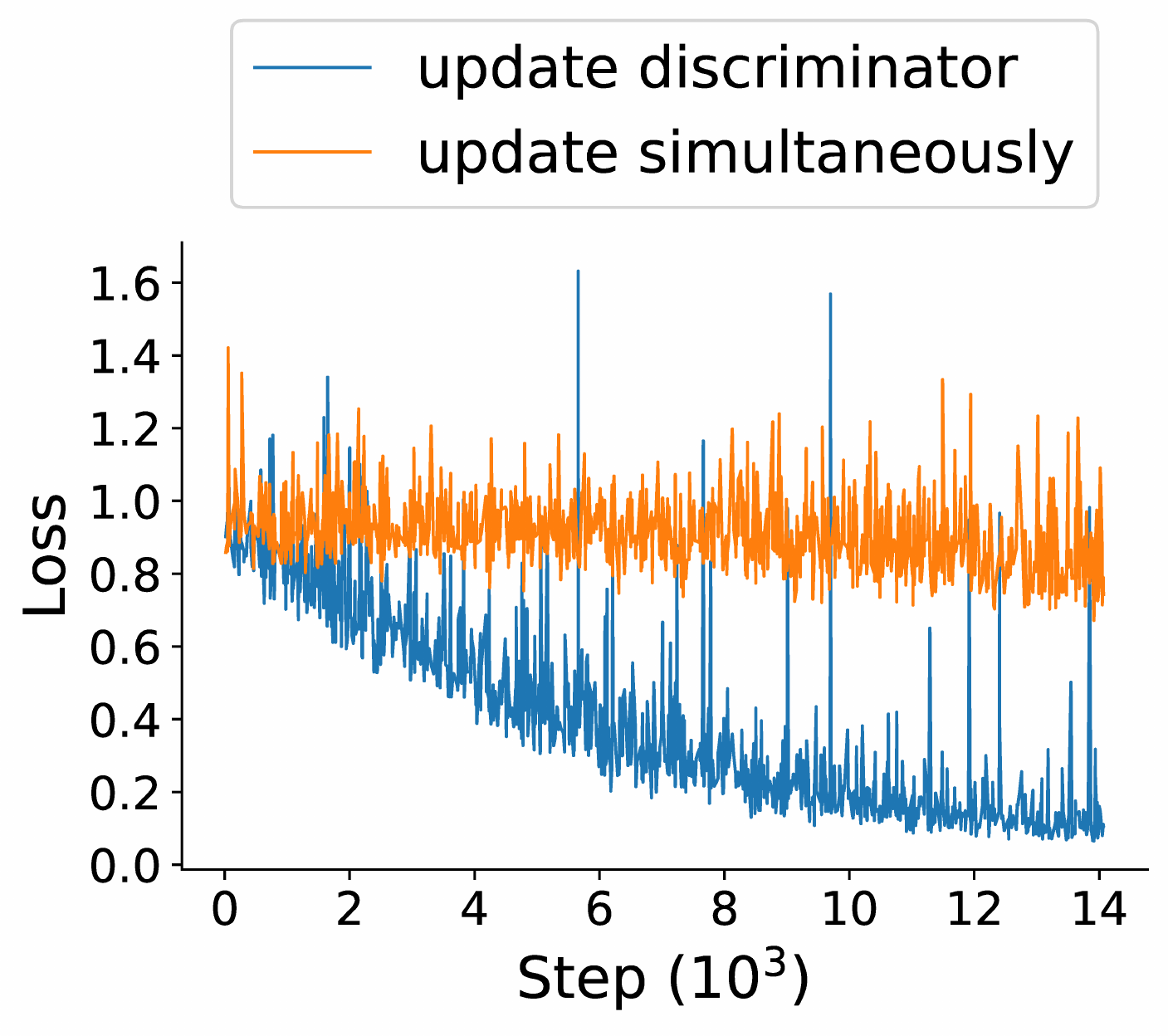}
    \includegraphics[width=0.49\columnwidth]{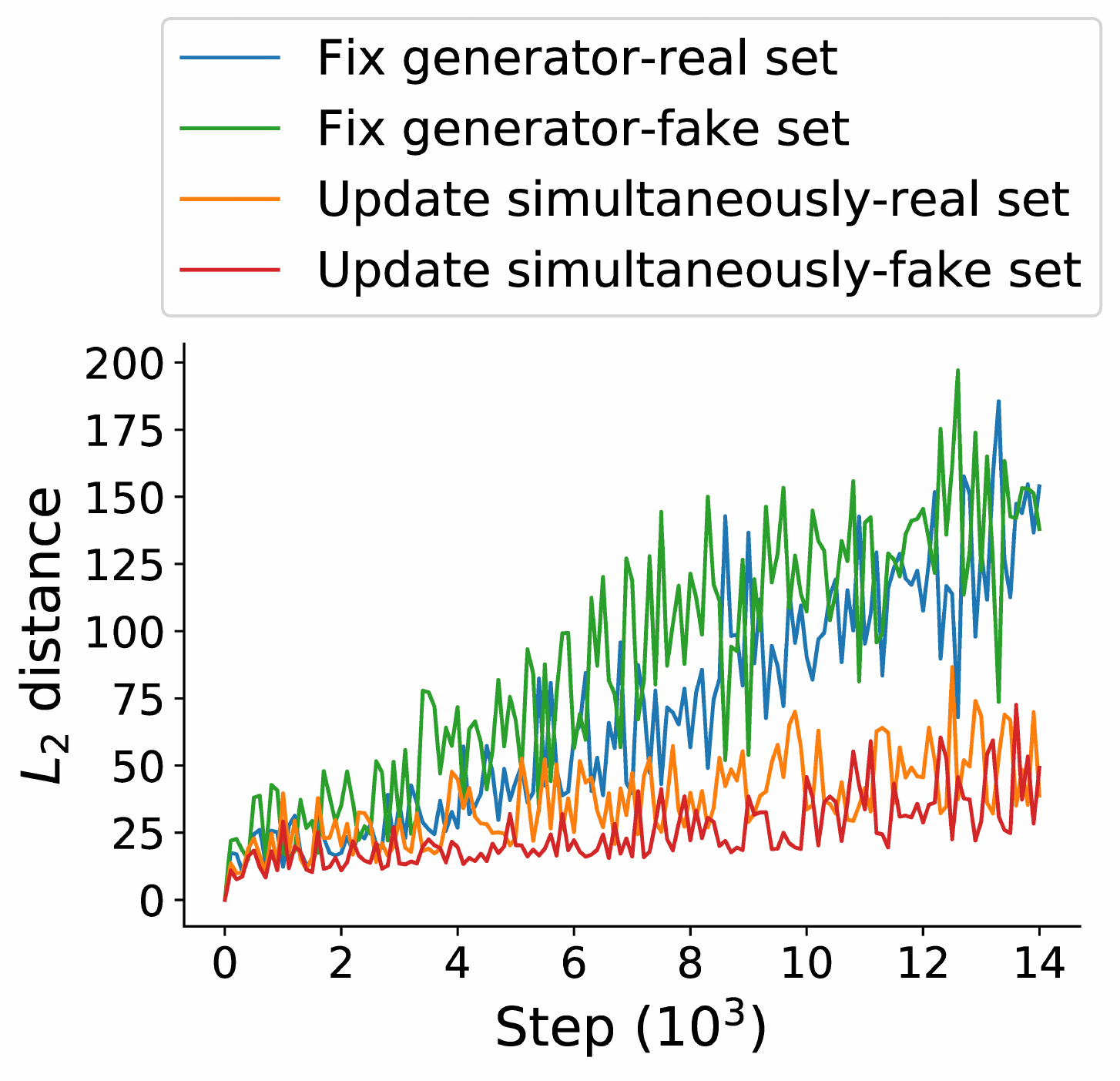}
    \caption{\textbf{ICR on MNIST:} We train a GAN on MNIST until we reach a \emph{checkpoint} where it produces good images. 
    (First image:) We fix the generator and only train the discriminator, observing that it can reach near-zero loss. When instead training generator and discriminator jointly, the loss stays stable. 
    (Second Image:) When trained individually, the discriminator moves significantly slower slower when trained jointly with the generator, as measured by its output on a set of thousand reference images.}
    \label{fig:overtraining_1}
\end{figure}

\section{How ICR lets GANs generate}
\textbf{An (hypo)thesis:}
In the example in the last section, the checkpoint producing good images was stabilized by ICR.
However, we have not yet given a reason why points stabilized by ICR should have better generators, in general.
For GANs to produce visually plausible images, there has to be some correspondence between the training of neural networks and human visual perception.
Since learning and generalization are poorly understood even for ordinary neural network classifiers, we can not avoid making an assumption on the nature of this relationship.
This section relies on the following hypothesis.

\textbf{Hypothesis} How quickly the discriminator can pick up on an imperfection of the generator is correlated with the visual prominence of said imperfection.

\begin{figure}
\centering 
\includegraphics[width=0.9\columnwidth]{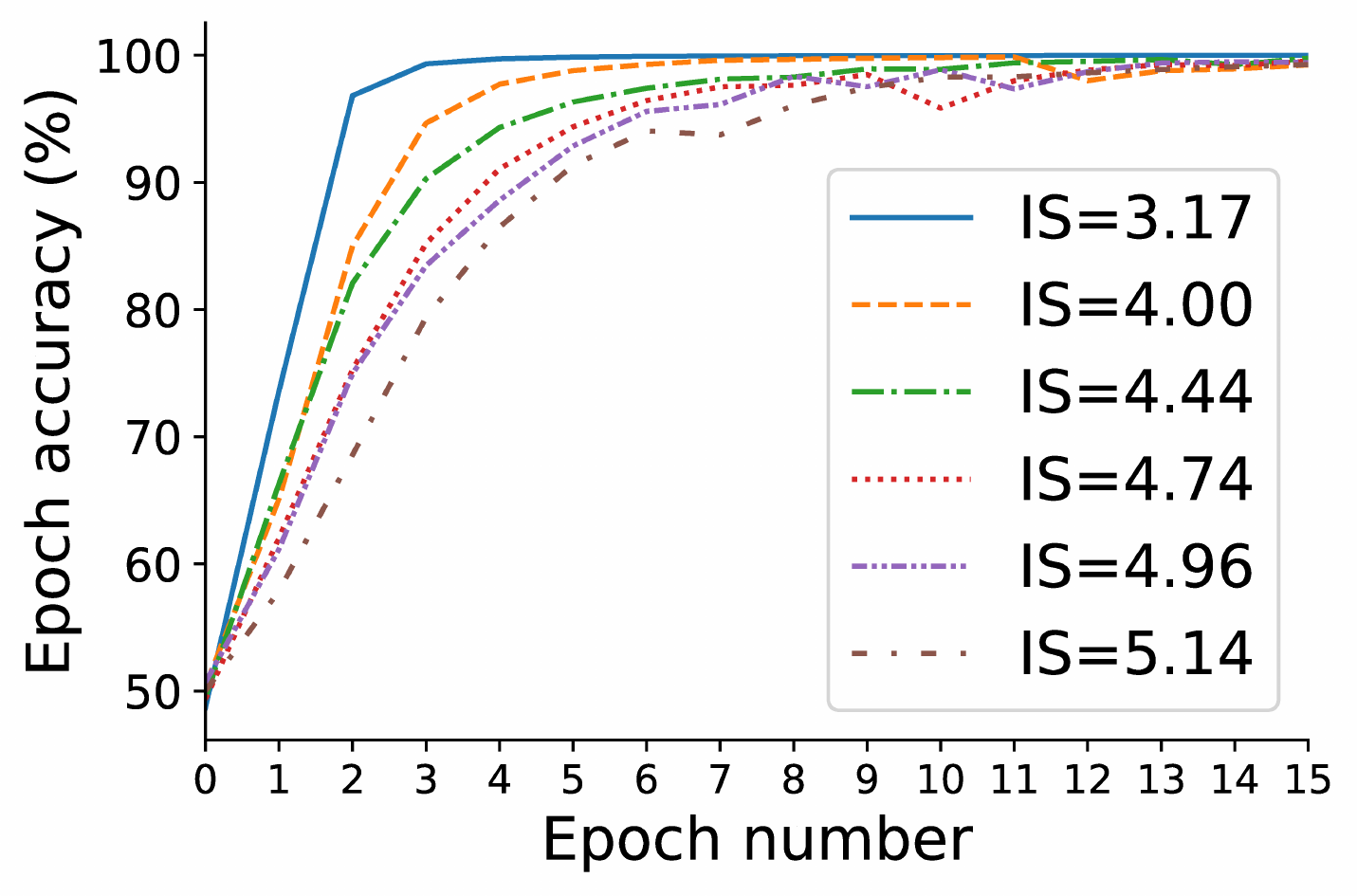}
\caption{\label{fig:pretrain} By prematurely stopping the training process, we obtain generators of different image-quality on CIFAR10  (higher inception score (IS) reflects better image quality). 
We then train a new discriminator against this \emph{fixed} generator and measure how quickly it increases its classification performance. 
We use a model trained on the 10-class classification task as starting point for the discriminator to prevent the initial phase of training from polluting the measurements.
While all discriminators achieve near-perfect accuracy eventually, the \emph{rate} of improvement is inversely correlated to inception score of the generator.}
\end{figure}

It is common intuition in training neural network classifiers that more visually obvious patterns are learned in fewer iterations and from less data. 
It is also in line with the \emph{coherent gradient hypothesis} of \citet{chatterjee2020coherent} that explains generalization performance of neural networks with the fact that systematic patterns in the data generate more coherent gradients and are therefore learned faster.
While a thorough verification of the hypothesis is beyond the scope of this work we provide some empirical evidence in Figure~\ref{fig:pretrain}.

This section argues for the following thesis.

\textbf{Thesis:} ICR selectively stabilizes generators for which the discriminator can only improve its loss  \emph{slowly}.
By the hypothesis, these generators will produce high quality samples.

\begin{figure}
    \centering
    \includegraphics[width=0.9\columnwidth]{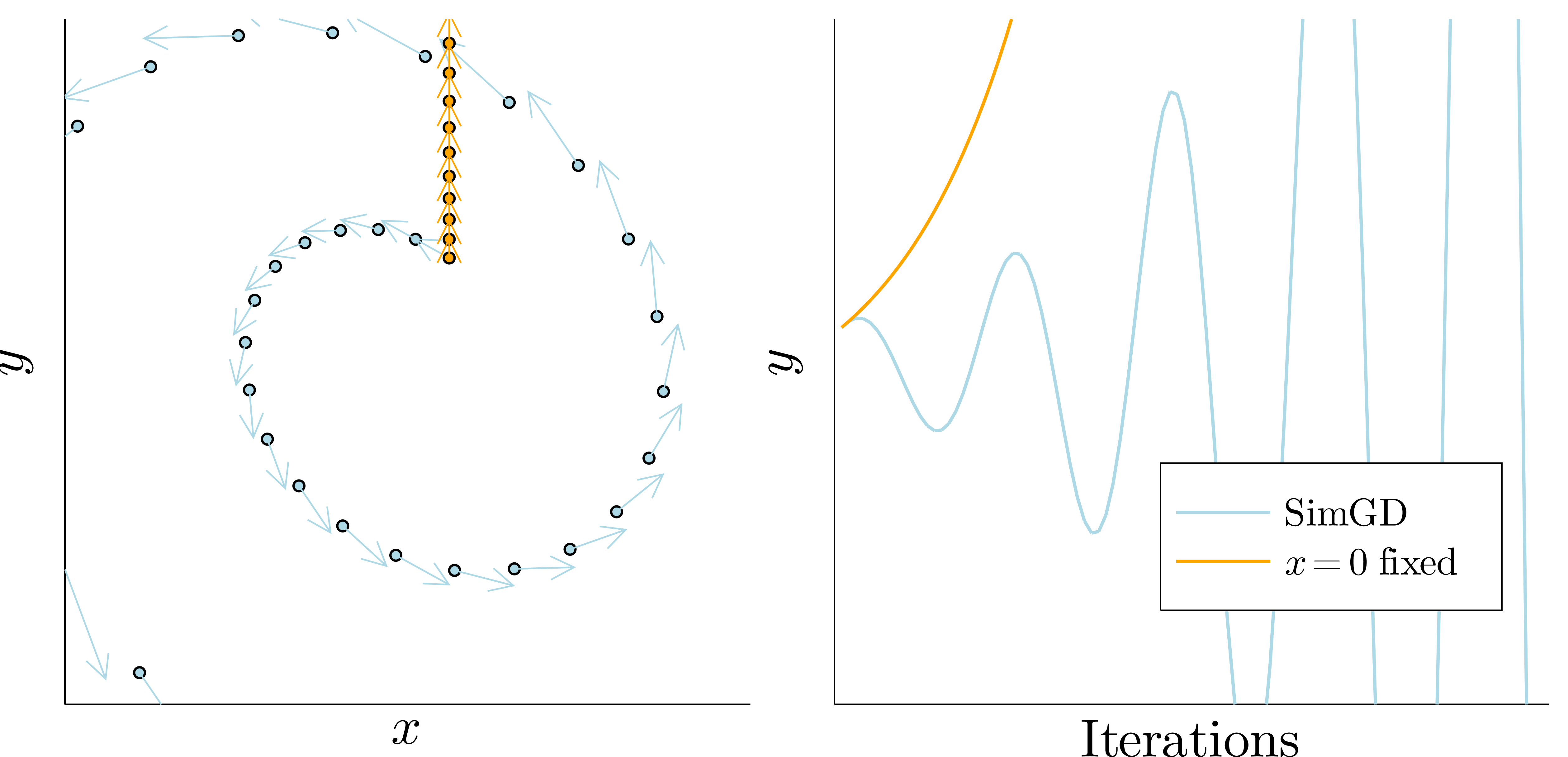}
    \includegraphics[width=0.9\columnwidth]{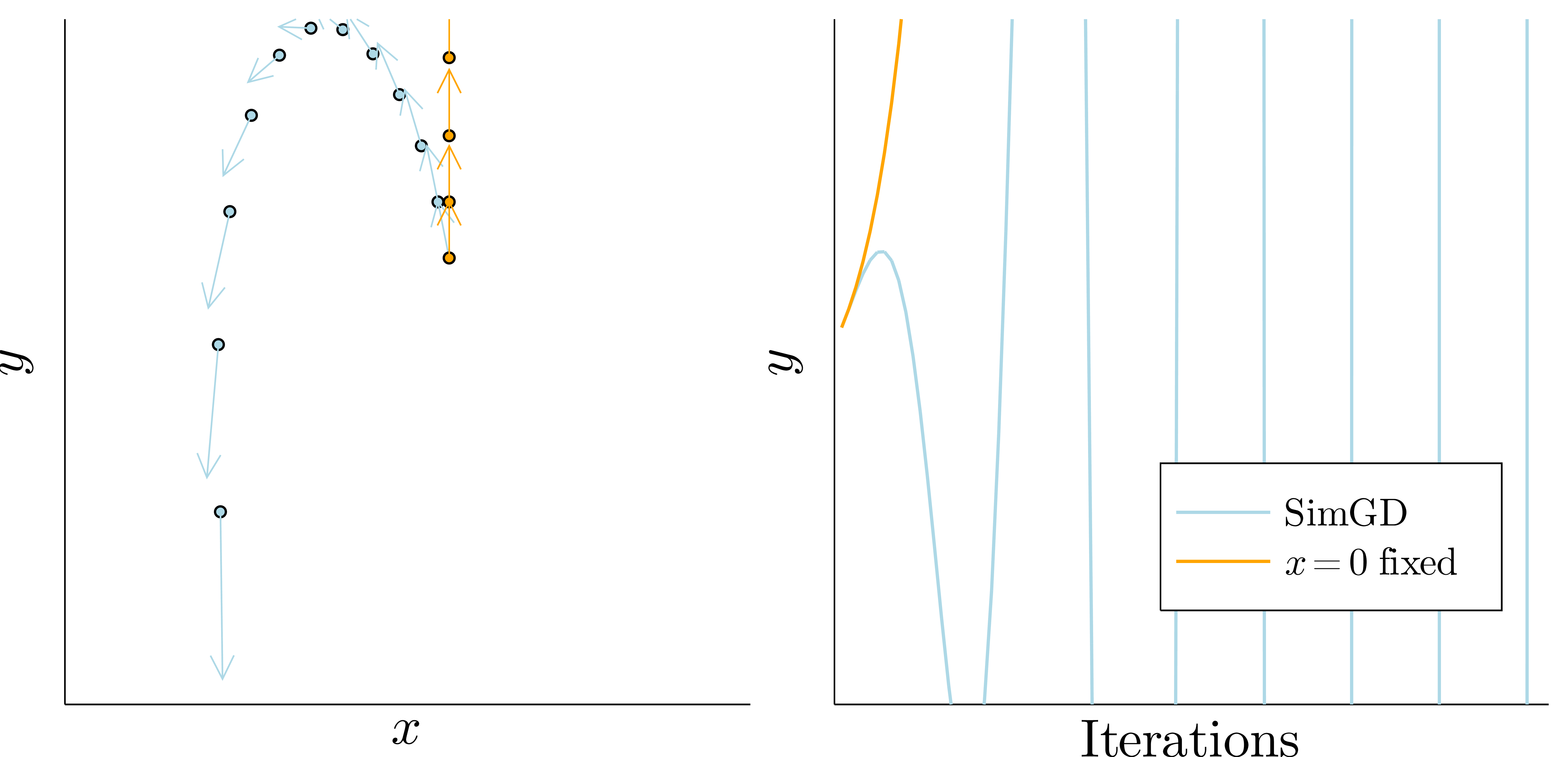}
    \caption{\textbf{ICR depends on speed of learning:} When changing the learning rates to $\left(\eta_{x}, \eta_{y}\right) = \left(0.03, 0.03\right)$ (top) or $\left(\eta_{x}, \eta_{y}\right) = \left(0.01, 0.09\right)$ (bottom), SimGD diverges.} 
    \label{fig:step_sizes}
\end{figure}

\textbf{An argument in the quadratic case:}
We begin with the quadratic problem in Equation~\ref{eqn:basicICR} and model the different speeds of learning of the two agents by changing their step sizes $\eta_x$ and $\eta_y$. 
In Figure~\ref{fig:step_sizes} we see that for $\left(\eta_{x}, \eta_{y}\right) = (0.03, 0.03)$ the two agents slowly diverge to infinity and for $\left(\eta_{x}, \eta_{y}\right) = \left(0.01, 0.09\right)$, divergence occurs rapidly.
In general, stable points $\bar{x}$ of an iteration $(x_{k + 1} = x_{k} + F\left(x_{k}\right)$ are characterized by \textbf{(1):} $F\left(\bar{x}\right) = 0$ and \textbf{(2)} $D_{x}F\left(\bar{x}\right)$ having spectral radius smaller than one \citep{mescheder2017numerics}[Proposition 3].
For SimGD applied to a zero sum game with the loss of $x$ given by $f$, these are points with vanishing gradients such that 
\begin{equation*}
\Id - M 
\coloneqq
\Id - 
\begin{pmatrix}
\eta_x D_{xx}^2 f & \eta_{x} D_{xy}^2 f \\
- \eta_y D_{yx}^2 f & -\eta_{y} D_{yy}^2 f 
\end{pmatrix}
\end{equation*}
has spectral radius smaller than one.
For univariate $x$, $y$ we can set $a \coloneqq D_{xx}^2f$, $b \coloneqq D_{xy}^2f$, and $c \coloneqq D_{yy}^2f$ and compute the characteristic polynomial of $M$ as 
\begin{equation*}
    p(\lambda) = \lambda^2 - (\eta_x a - \eta_y c) \lambda + (- \eta_x \eta_y a c + \eta_x \eta_y b^2).
\end{equation*}
For $\eta_{x} a > \eta_{y}c$  and $ \eta_x \eta_y b^2 > \eta_x \eta_y a c$ the solutions of this equation have positive real part and therefore the eigenvalues of $M$ have positive real part.
By multiplying $\eta_{x}$ and $\eta_{y}$ by a small enough factor we can obtain a spectral radius smaller than one (c. f. \citet{mazumdar2018convergence}).
Thus, a small enough $\eta_{y}$ and large enough mixed derivative $b$ can ensure convergence even for positive $c$.

If we think of the maximizing player as the discriminator, slow learning (modelled by small $\eta_{y}$) is correlated to good images produced by the generator.
Thus, in this interpretation, a good generator leads to ICR stabilizing the point $(0,0)$ more strongly.

\begin{figure}
    \centering
    \includegraphics[scale=0.18]{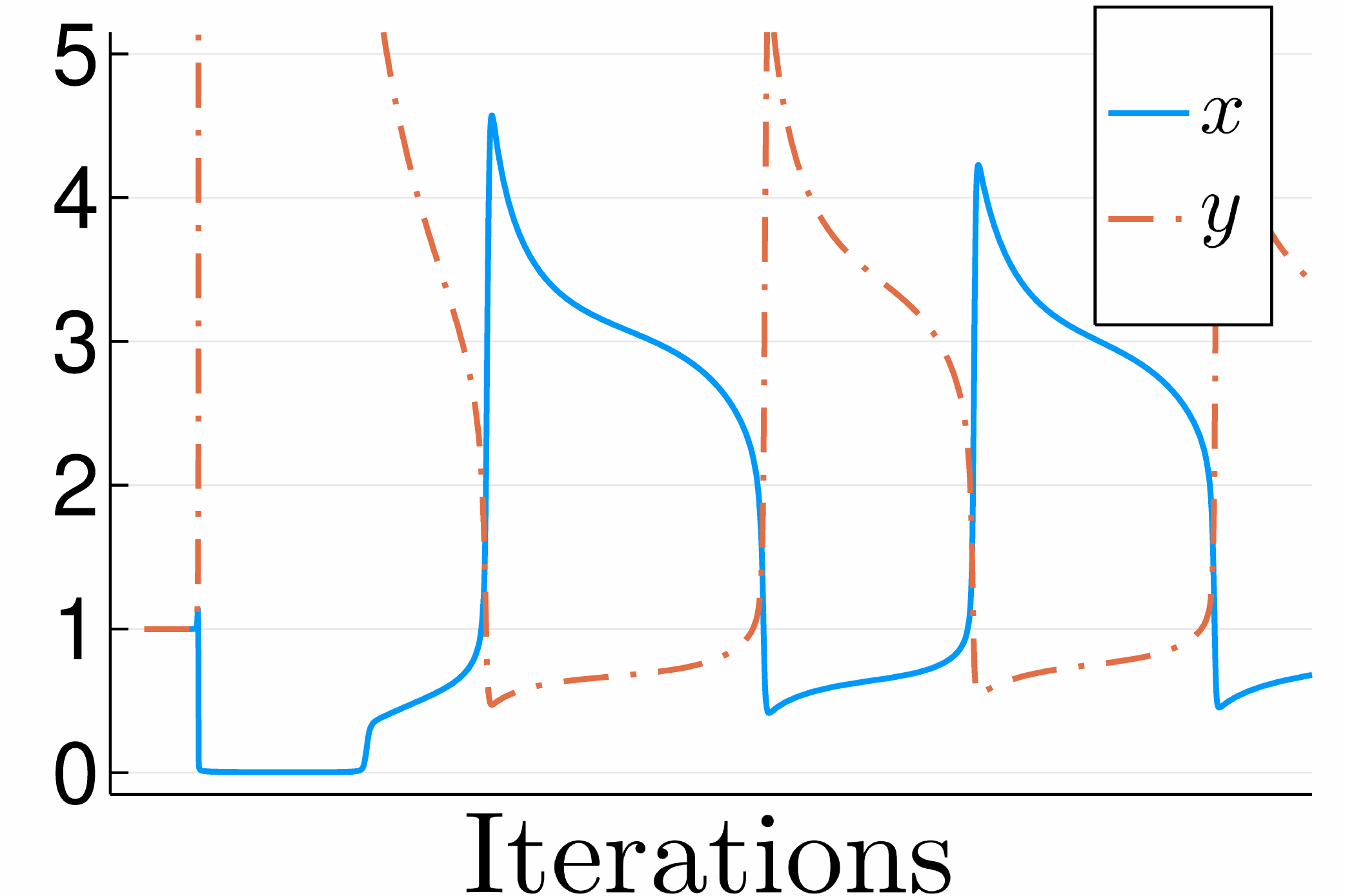}
    \includegraphics[scale=0.18]{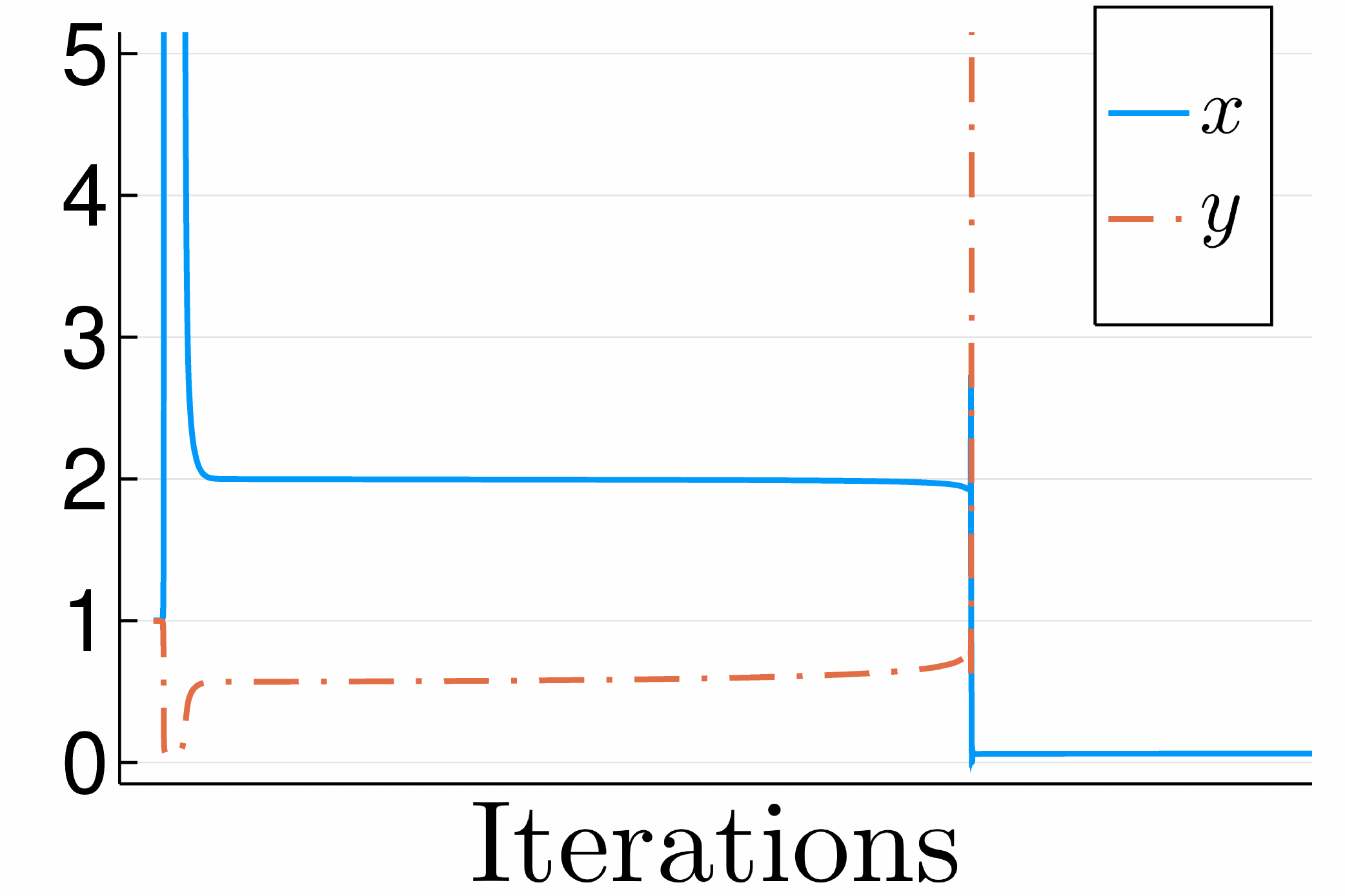}
    \includegraphics[scale=0.25]{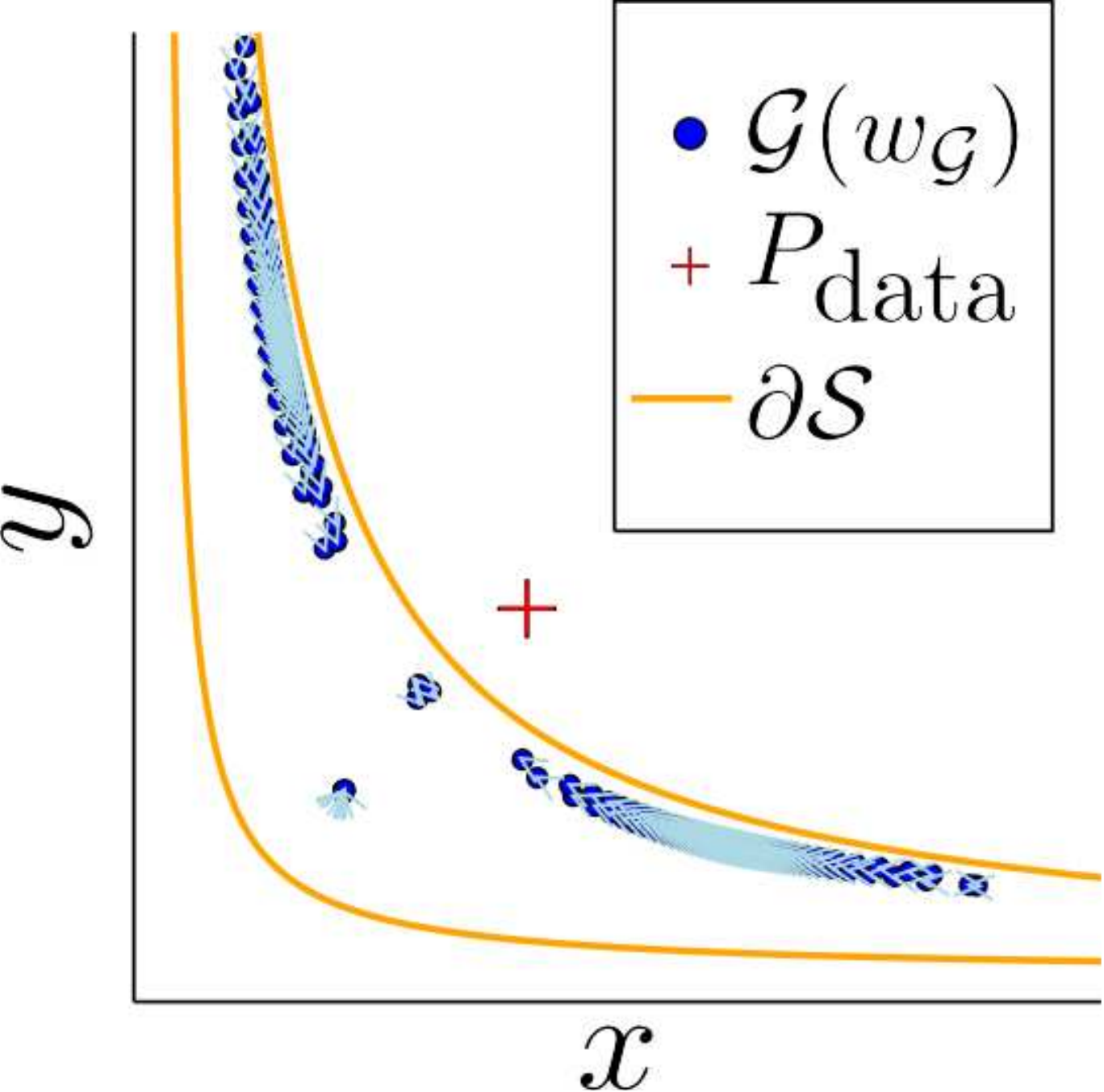}
    \includegraphics[scale=0.25]{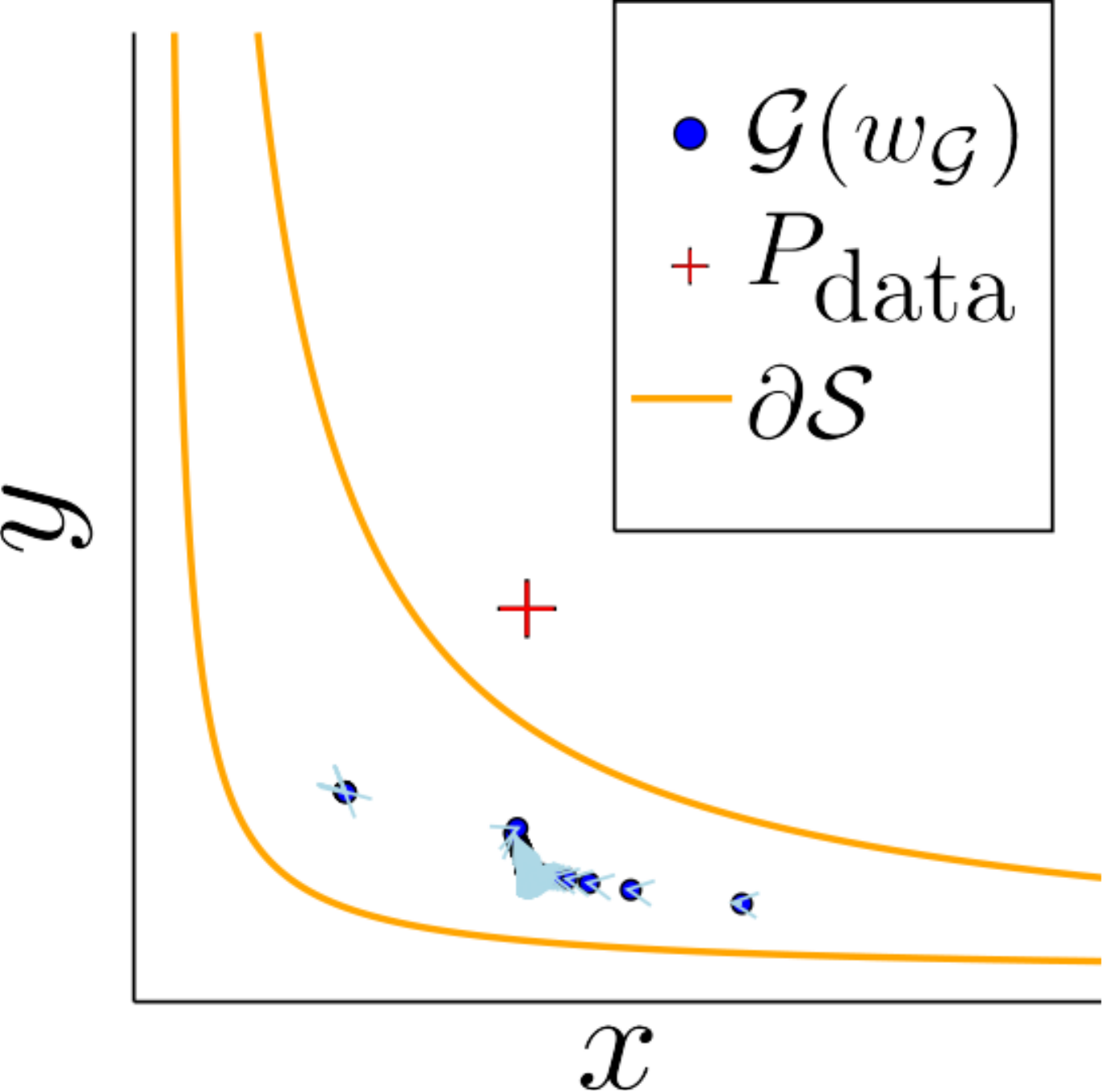}
    \caption{\textbf{Approximate projection via adversarial training:} On the left column, discriminator picks up on errors in the $x$- and $y$-direction equally quickly. Therefore, the generator tries to satisfy the criteria alternatingly, leading to a cyclic pattern. In the right column, the discriminator picks up on errors in the $x$-direction much more quickly. This causes the generator to try to stay accurate in the $x$-direction.}
    \label{fig:toyGAN}
\end{figure}

\textbf{Adversarial training as projection:}
Surprisingly, ICR allows us to compute a projection with respect to the perceptual distance of a neural network, without quantifying this distance explicitly.
Let us consider the following example.
We construct a \emph{generator} $\G$ that maps its $28$ weights to a bivariate output. 
This nonlinear map is modelled as a tiny neural network with two hidden layers, with the final layer restricting the output to the set $\mathcal{S} \coloneqq \left\{ (e^{s+t}, e^{s-t}) \middle| s \in \left[-\frac{1}{2}, \frac{1}{2}\right], t \in \Reals \right\} \subset \Reals^2$.
We think of this as mapping a set of weights to a generative model that is characterized by only two parameters.
In this parameterization, we assume that the target distribution is represented by the point $\Pd = (2,2)$. 
Importantly, as shown in Figure~\ref{fig:toyGAN}, there is no set of weights that allow the generator to output \emph{exactly} $\Pd$.
This is to model the fact that in general, the generator will not be able to exactly reproduce the target distribution.
We construct a \emph{discriminator} $\D$ that maps a generative model (a pair of real numbers) and a set of 28 weights to a real number, by a small densely connected neural network. 

We want to model the difference in visual prominence of the two components of $\Pd$. To this end, we assume that before before being passed to the discriminator, $\G$ and $\Pd$ are rescaled by a diagonal matrix $\eta \in \Reals^{2 \times 2}$.
Thus, $\eta$ determines the relative size of the gradients of $\D$ of the first and second components of the input data.
This models the hypothesis that a real discriminator will pick up more quickly on visually prominent features. 
Importantly, we assume $\eta$ to be unknown, since we do not have access to a metric measuring "visual similarity to a neural network".\\

We will now show how adversarial training can be used to approximate a projection with respect to $\eta$, without knowing $\eta$.
We use the loss
\begin{equation}
   \min \limits_{w_{\G} \in \Reals^{28}} \max \limits_{w_{\D} \in \Reals^{28}} \D\left(\eta \Pd, w_{\D} \right) - \D\left(\eta \G\left(w_{\G}\right), w_{\D}\right)
\end{equation}
and train the two networks using simultaneous gradient descent.
For $\eta$ equal to the identity, we see oscillatory training behavior as $\G$ tries be accurate first in one, then the other direction.
If we instead use $\eta = \left(\begin{smallmatrix} 1 & 0\\ 0  & 10^{-2} \end{smallmatrix}\right)$, we are modelling the first component as being more visually prominent.
Instead of the oscillatory patterns from before, we observe long periods where the value of the first component of $\G\left(w_{\G}\right)$ is equal to the first component of $\Pd$ (see Figure~\ref{fig:toyGAN}). 
Without knowing $\eta$, we have approximated the projection of $\Pd$ onto $\mathcal{S}$ with respect to the metric given by $(x,y) \mapsto \|\eta (x,y)\|$.
To do so, we used the fact that this point is subject to the slowest learning discriminator, and thus the strongest ICR.

We believe that GANs use the same mechanism to compute generators that are close to the true data in the perceptual distance of the discriminator, which in turn acts as a proxy for the perceptual distance of humans.

\section{Competitive gradient descent amplifies ICR}

\textbf{How to strengthen ICR:}
We have provided evidence that GANs' ability to generate visually plausible images can be explained by ICR selectively stabilizing good generators.
It is well known that GANs often exhibit unstable training behavior, which is mirrored by the observations in Figures~\ref{fig:cycling}, ~\ref{fig:overtraining_1} and~\ref{fig:toyGAN} that ICR often only leads to weak, temporary stability.
Thus, it would be desirable to find algorithms that induce stronger ICR than SimGD.
To this end, we will find a game-theoretic point of view useful.

\textbf{Cooperation in a zero-sum game?}
As discussed in the last section, ICR can stabilize solutions that are locally suboptimal for at least one of the players.
Since we did not model either of the two players as altruistic, this behavior may seem puzzling.
It is likely for this reason that ICR has mostly been seen as a flaw, rather than a feature of SimGD.

\textbf{Convergence by competition:}
The quadratic example in Equation~\eqref{eqn:basicICR} shows that the bilinear term $xy$ is crucial for the presence of ICR. Otherwise,  SimGD   reduces to each player moving independently according to gradient descent.
In fact, the strength of ICR decreases rapidly as $\left|\alpha\right|$ and $\left|\beta\right|$ diverge to infinity.

The mixed term $xy$ models the ability of each player to retaliate against actions of the other player.
In the case of $\beta < 0$, as the maximizing player $y$ moves to plus infinity in order to maximize its reward, it becomes a locally optimal strategy for the minimizing player $x$ to move towards negative infinity in order to minimize the dominant term $xy$.
If $|\beta| \ll 1$ it is favorable for the maximizing player to move back towards zero in order to maximize the dominant term $xy$.
The reason for the maximizing player to stay in the sub-optimal point $y=0$ (the \emph{maximizer} of its loss, for $x = 0$) is the that minimizing player can use the mixed term $xy$ to punish every move of $y$ with a counterattack.
Thus, the need to avoid counterattacks justifies the seemingly sub-optimal decision of the maximizing player to stay in $y=0$.

\begin{figure}
    \centering
    \includegraphics[width=0.49\columnwidth]{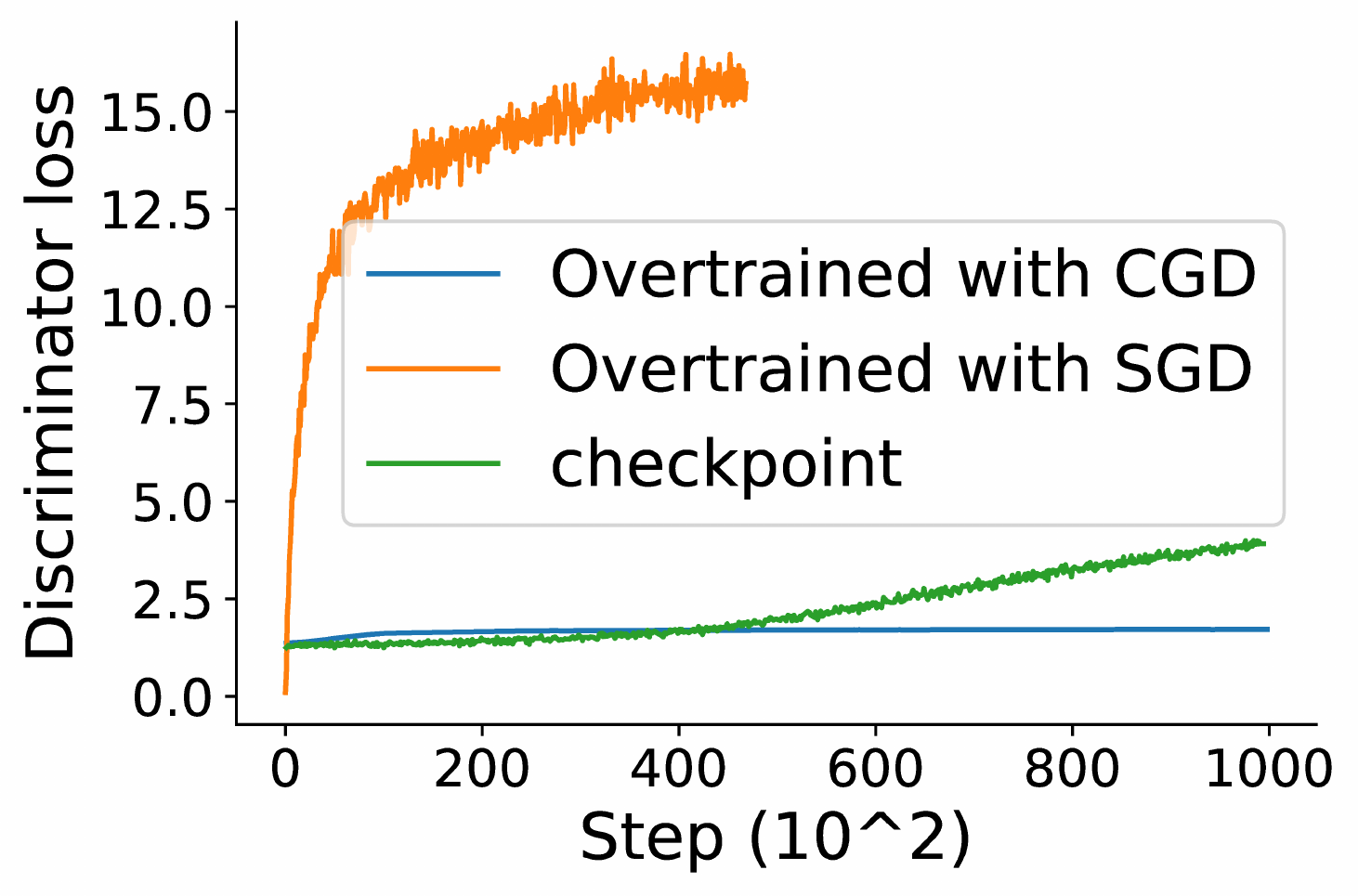}
    \includegraphics[width=0.49\columnwidth]{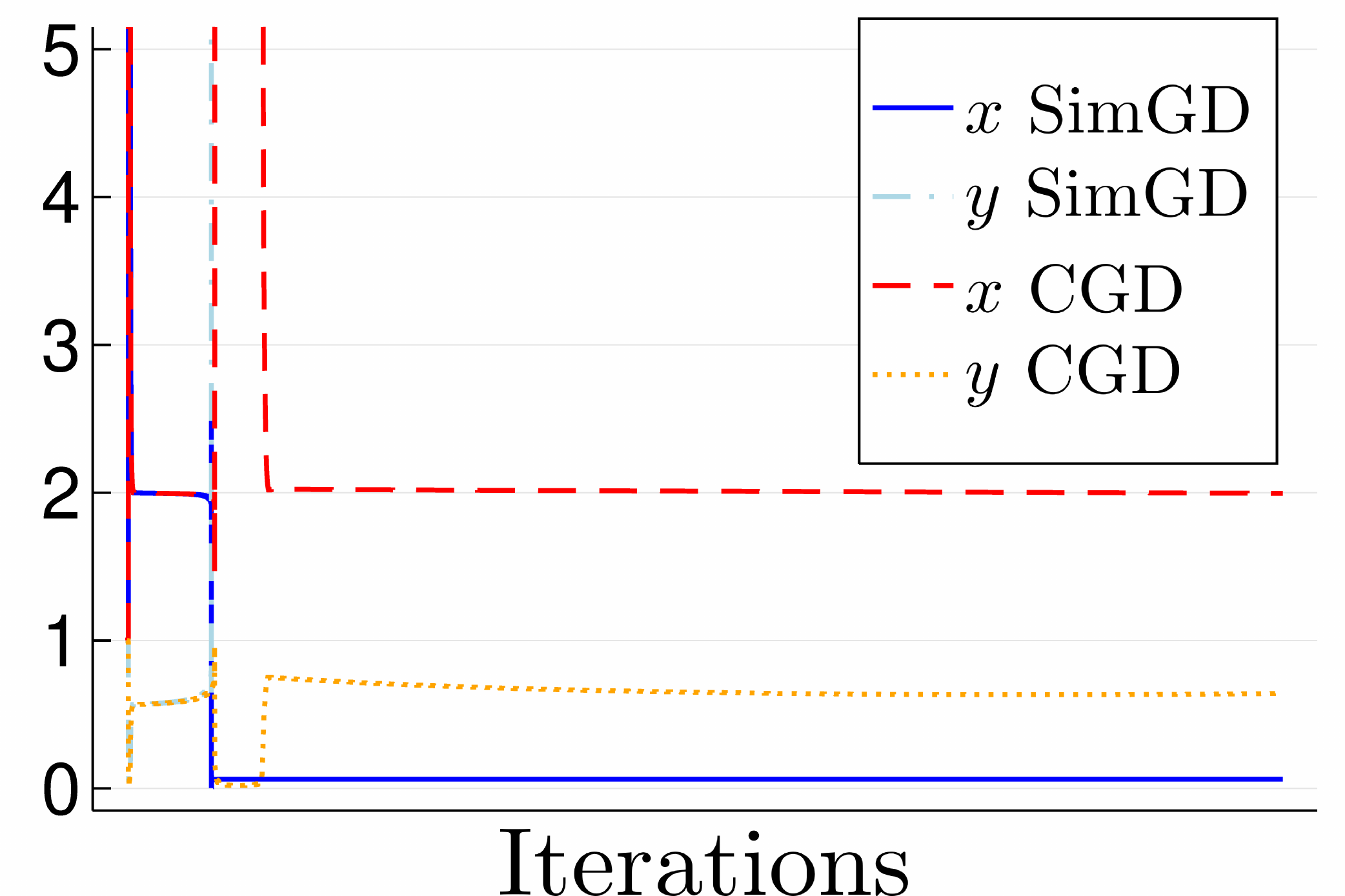}
    \caption{\textbf{ICR and opponent-awareness:} When training the generator for just a few iterations against the over-trained discriminator of Figure~\ref{fig:overtraining_1}, the discriminator loss increases rapidly. When attempting to over-train with CGD instead of Adam, the resulting discriminator is even more robust. 
    Similarly, CGD is able to significantly increase the duration for which the generator stays accurate in the (more important) $x$-direction in Figure~\ref{fig:toyGAN}.}
    \label{fig:overtraining_2}
\end{figure}

\textbf{The generator strikes back!}
This phenomenon is also present in the example of Figure~\ref{fig:overtraining_1}.
Consider the checkpoint generator from Figure~\ref{fig:overtraining_1} and the over-trained discriminator that achieves near perfect score against the discriminator.
As we can see in Figure~\ref{fig:overtraining_2}, training the generator while keeping the over-trained discriminator fixed leads to a rapidly increasing discriminator loss.
The over-trained discriminator has become vulnerable to counterattack by the generator!
If instead the generator is trained against the checkpoint discriminator, the loss increases only slowly.
Thus, ICR can be interpreted as the discriminator trying to avoid counterattack by the generator.

\textbf{Agent modelling for stronger ICR:}
The update $(x,y)$ of SimGD applied to the loss function $f$ can be interpreted as the two players solving, at each step, the local optimization problem
\begin{equation*}
    \min \limits_x x^{\top}\nabla_x f(x_k, y_k) + \frac{\|x\|^2}{2\eta}, \ \ \ \  
    \max \limits_y y^{\top} \nabla_y f(x_k, y_k) - \frac{\|y\|^2}{2\eta} 
\end{equation*}
The terms $x^{\top}\nabla_x f(x_k, y_k)$, $y^{\top} \nabla_y f(x_k, y_k)$ express the \emph{belief} about the loss associated to different actions, based on local information.
The quadratic regularization terms express their their \emph{uncertainty} about these beliefs, letting them avoid extreme actions (large steps).
However, $y$ ($x$) does not appear in the local optimization problem of $x$ ($y$).
Thus, the two players are not taking the presence of their opponent into account when choosing their actions.
Accordingly, ICR arises only because of the players reaction to, rather than anticipation of each others actions.
We propose to strengthen ICR by using local optimization problems that model the players' \emph{anticipation} of each other's action. 

\textbf{Competitive gradient descent:}
The updates of competitive gradient descent (CGD) \citep{schaefer2019competitive} are obtained as Nash equilibria of the local game
	\begin{align*}
	&\min \limits_x x^{\top}\nabla_x f(x_k, y_k) + x^{\top} [D_{xy}f(x_k, y_k))] y +   \frac{\|x\|^2}{2\eta}, \\
	&\max \limits_y y^{\top} \nabla_y f(x_k, y_k) + y^{\top} [D_{yx}f(x_k, y_k))] x - \frac{\|y\|^2}{2\eta}.
	\end{align*}
Under CGD, the players are aware of each other's presence at every step, since the mixed Hessian $x^{\top} [D_{xy}f(x_k,y_k)]y$ informs each player, how the simultaneous actions of the other player could affect the loss incurred due to their own action.
This element of anticipation strengthens ICR, as indicated by the convergence results provided by \citet{schaefer2019competitive}.
Providing additional evidence, we see in Figure~\ref{fig:overtraining_2} that attempting to over-train the discriminator using CGD leads to a discriminator that is even more robust than the checkpoint discriminator.
Applying CGD to the example of Figure~\ref{fig:toyGAN} also increases the stability of the approximate projection of $\Pd$ onto $\mathcal{S}$ according to the metric implicit in the discriminator.
These results suggest to use CGD to strengthen ICR in GAN training, which we will investigate in the next section.
We also expect methods such as LOLA \citep{foerster2018learning} or SGA \citep{balduzzi2018mechanics,gemp2018global} to strengthen ICR, but a detailed comparison is beyond the scope of this work.

\section{Empirical study on CIFAR10}
\textbf{Experimental setup:} Based on the last section, CGD strengthens the effects of ICR and should therefore improve GAN performance.
We will now investigate this question empirically.
In order to make for a fair comparison with Adam, we combine CGD with a simple RMSprop-type heuristic to adjust learning rates, obtaining adaptive CGD (ACGD, see supplement for details).
As loss functions, we use the original GAN loss (OGAN) of \eqref{eqn:ogan} and the Wasserstein GAN  loss function (WGAN) 
given by
\begin{equation*}
    \label{eqn:minmaxganws}
    \min \limits_{\mathcal{G}} \max \limits_{\mathcal{D}} ~ \Expect_{x \sim \Pd} \left[\mathcal{D}(x)\right] - \Expect_{x \sim \Pg} \left[\mathcal{D}(x)\right].
\end{equation*}
When using Adam on OGAN, we stick to the common practice of replacing the generator loss by $\Expect_{x \sim \Pg}\left[ - \log\left(\D(x) \right]\right]$, as this has been found to improve training stability \citep{goodfellow2014generative,goodfellow2016deep}.
In order to be generous to existing methods, we use an existing architecture
intended for the use with WGAN gradient penalty \citep{gulrajani2017improved}.
As regularizers, we consider no regularization (NOREG), $\ell_2$ penalty on the discriminator with different weights (L2), Spectral normalization \citep{miyato2018spectral} on the discriminator (SN), or $1$-centered gradient penalty on the discriminator, following \cite{gulrajani2017improved} (GP).
Following the advice in \citep{goodfellow2016deep} we train generator and discriminator simultaneously, with the exception of WGAN-GP and Adam, for which we follow \citep{gulrajani2017improved} in making five discriminator updates per generator update.
We use the Pytorch implementation of inception score (IS) \citep{salimans2016improved} to compare generator quality.\footnote{The Pytorch implementation gives slightly different scores than Tensorflow. We report Tensorflow IS in the supplementary material showing that the relative performance is largely the same.
}
\begin{figure*}
	\centering
	\begin{minipage}{0.32\textwidth}
	\centering
	\includegraphics[width=\textwidth]{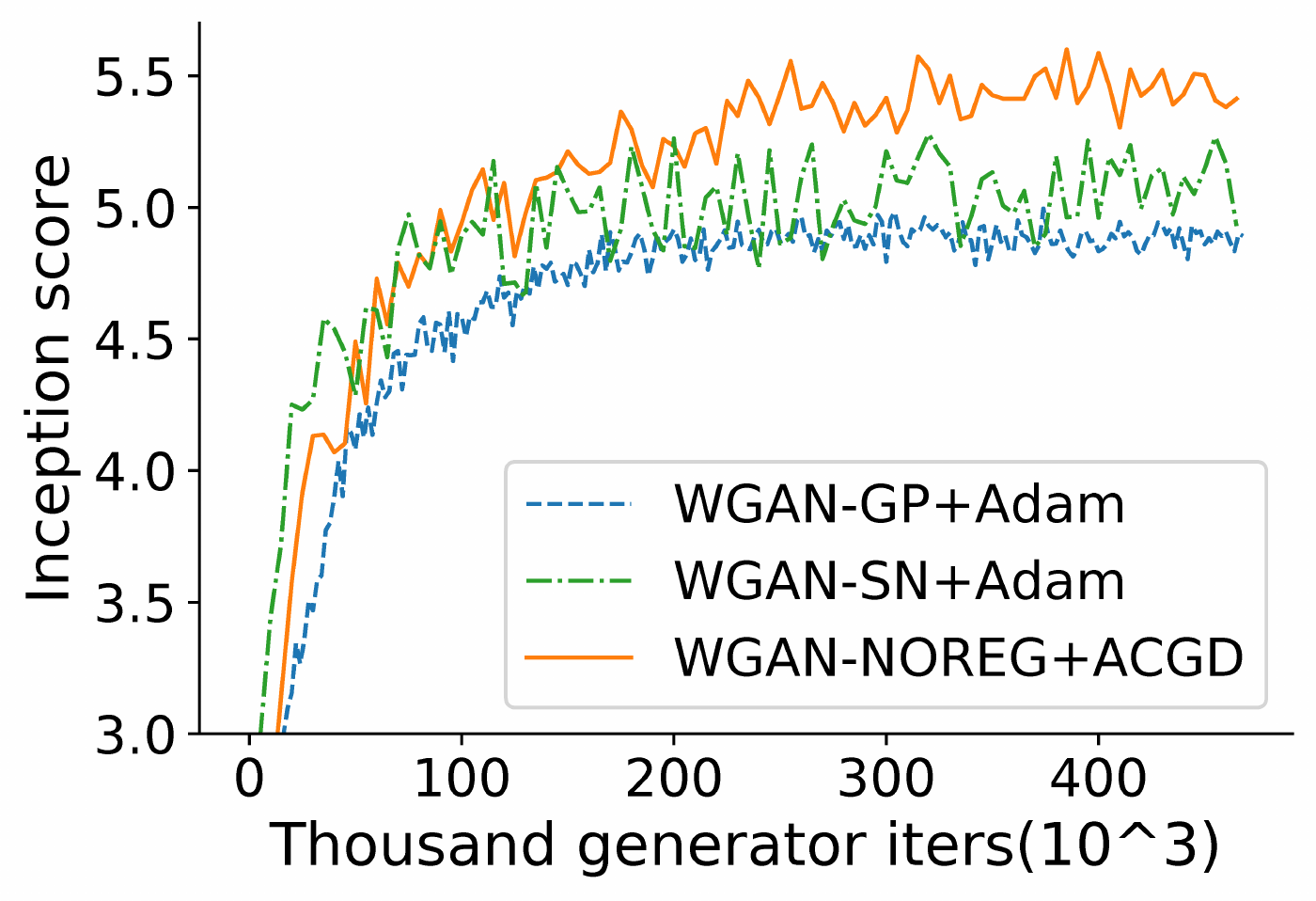}
	\end{minipage}
	\begin{minipage}{0.32\textwidth}
	\centering
	\includegraphics[width=\textwidth]{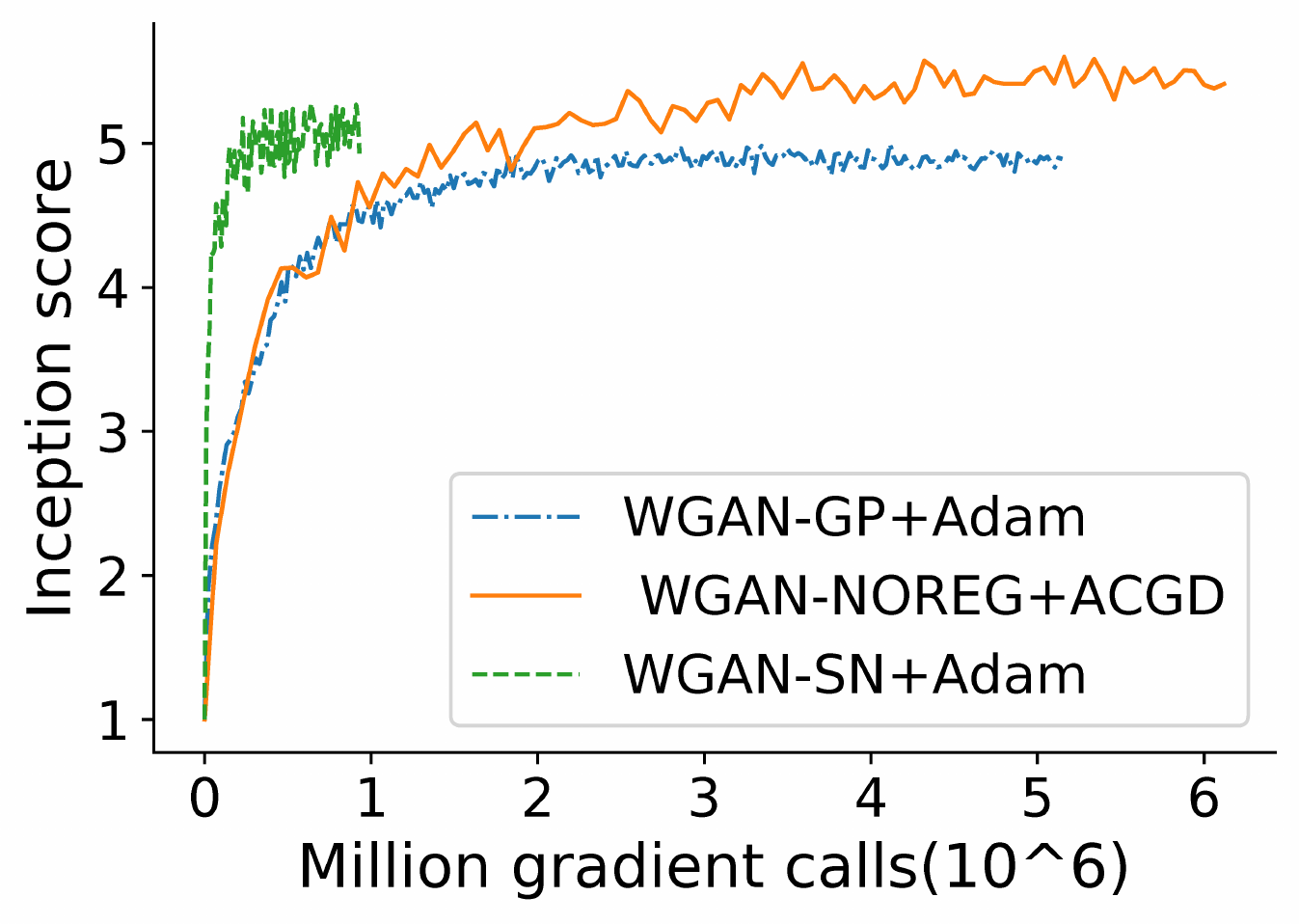}
	\end{minipage}
	\begin{minipage}{0.32\textwidth}
	\centering
	\includegraphics[width=0.70\textwidth]{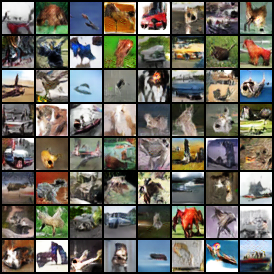}
	\end{minipage}
	
	\begin{minipage}{0.32\textwidth}
	\centering
	\includegraphics[width=\textwidth]{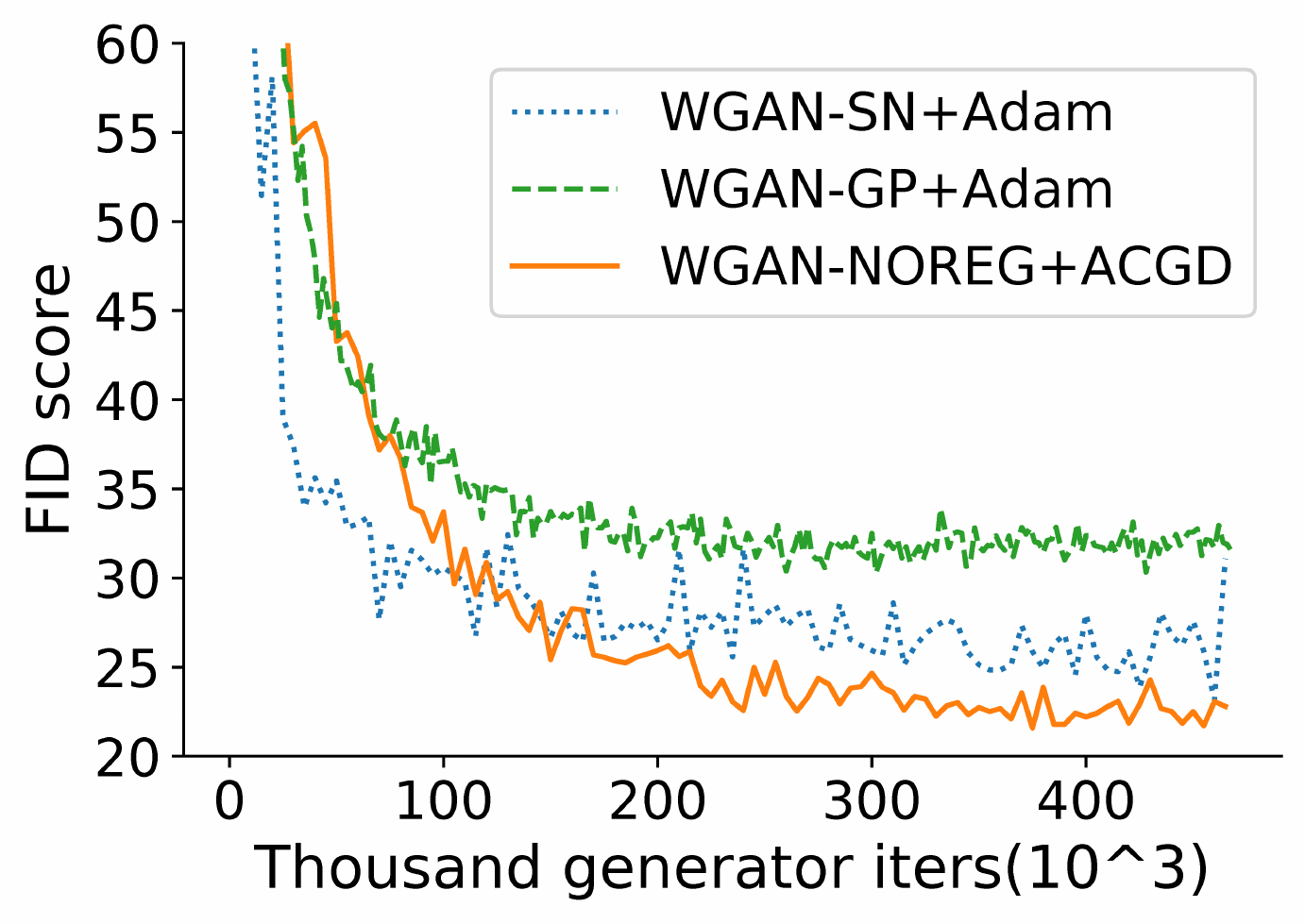}
	\end{minipage}
	\begin{minipage}{0.32\textwidth}
	\centering
	\includegraphics[width=\textwidth]{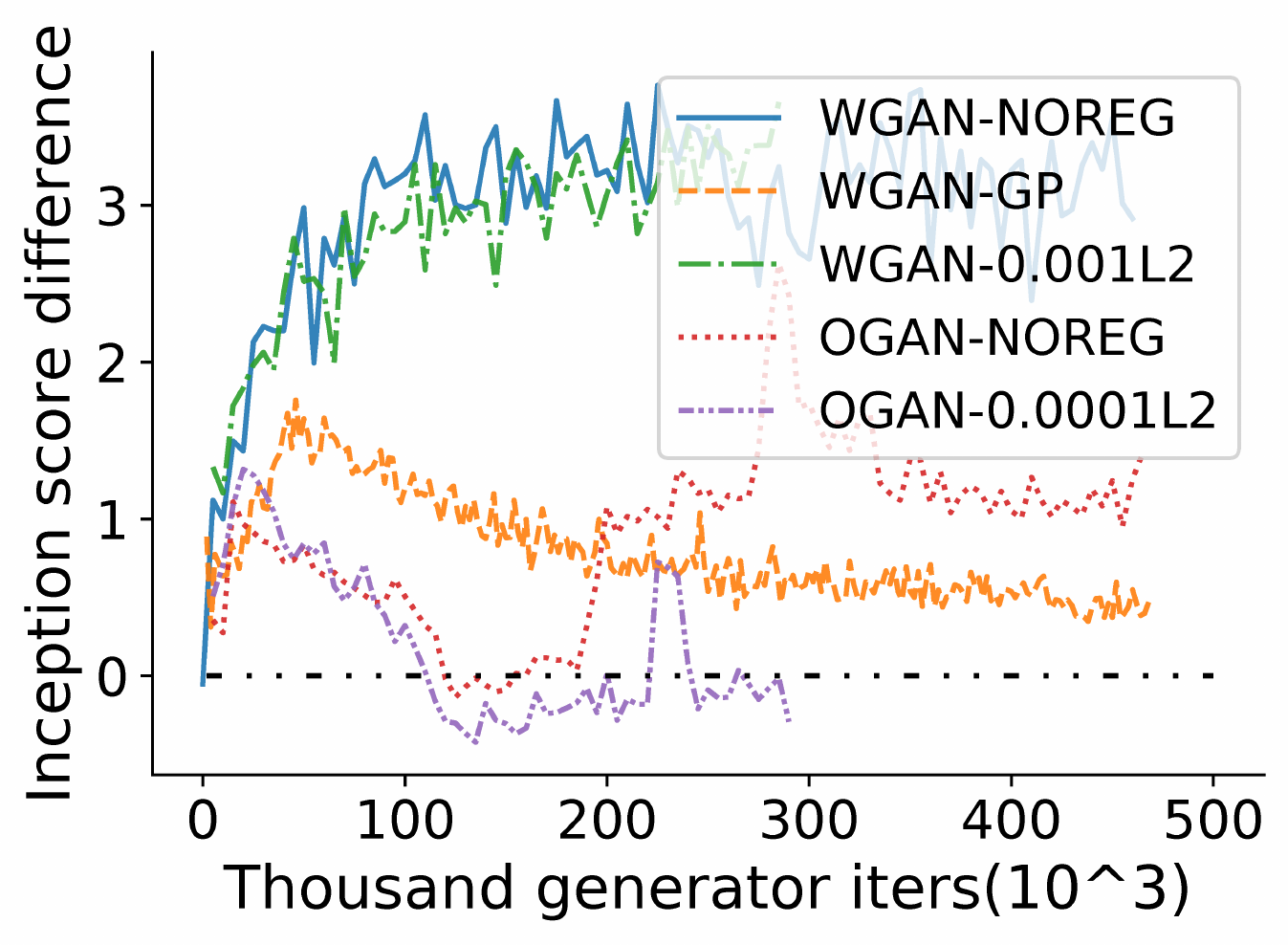}
	\end{minipage}
	\begin{minipage}{0.32\textwidth}
	\centering
	\includegraphics[width=\textwidth]{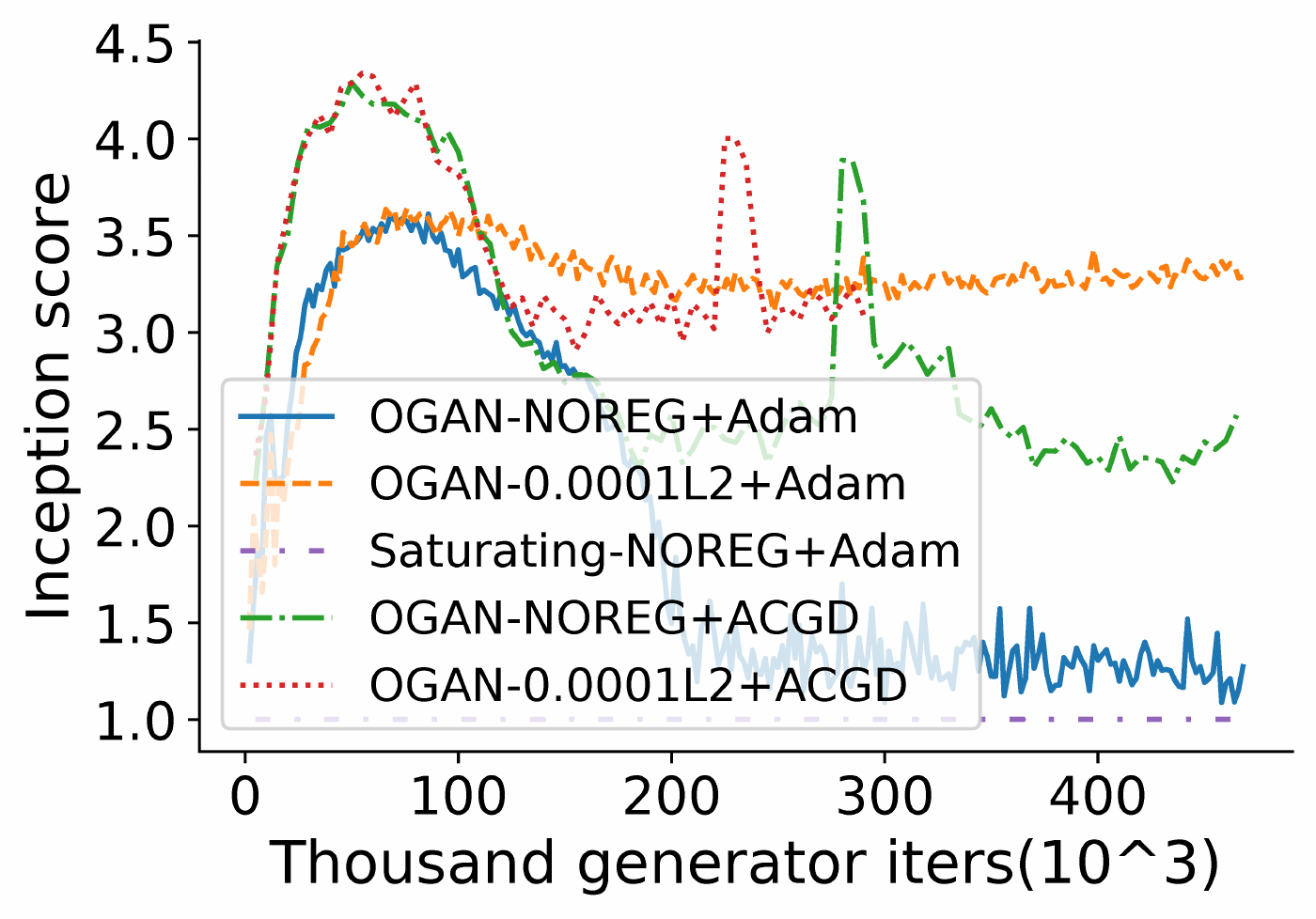}
	\end{minipage}
	\caption{We plot the inception score (IS) against the number of iterations (first panel) and gradient or Hessian-vector product computation (second panel). In the third panel we show final samples of WGAN trained with ACGD and without explicit regularization.
	In panel four, we compare measure image quality using the Frechet-inception-distance (FID, smaller is better). The results are consistent with those obtained using IS.
	In panel five, we plot the difference between inception scores between ACGD and Adam (positive values correspond to a larger score for ACGD) over different iterations and models. 
	The only cases where we observe nonconvergence of ACGD are OGAN without regularization or with weight decay of weight $0.0001$, as shown in the last panel. The inception score is however still higher than for the same model trained with Adam. 
	When using Adam on the original saturating GAN loss (which we used with ACGD), training breaks down completely.}
	\label{fig:summary}
\end{figure*}

\textbf{Experimental results:} We will now summarize our main experimental findings, (see Figure ~\ref{fig:summary}). 
\textbf{(1:)} When restricting our attention to the top performing models, we observe that the combination of ACGD with the WGAN loss and without any regularization achieves higher inception score than all other combinations tested.
\textbf{(2:)} The improvement obtained from training with ACGD persists when measuring image quality according to the Freched-inception-distance (FID) \citep{heusel2017gans}.
\textbf{(3:)} When comparing the number of gradient computations and Hessian-vector products, ACGD is significantly slower than WGAN loss with spectral normalization trained with ADAM, because of the iterative solution of the matrix inverse in ACGD's update rule. 
\textbf{(4:)} The only instance where we observe erratic behavior with ACGD is when using OGAN without regularization, or with a small $\ell_2$ penalty. However, ACGD still outperforms Adam on those cases. In particular training with Adam breaks down completely when using the original saturating loss (as we do for ACGD). 
\textbf{(5:)} When plotting the difference between the inception scores obtained by ACGD and Adam for the same model over the number of iterations, for all models, we observe that ACGD often performs significantly better, and hardly ever significantly worse.

Since CGD strengthens the effects of ICR, the performance improvements obtained with CGD provide further evidence that ICR is a key factor to GAN performance.

\section{Conclusion and outlook}
In this work, we have pointed out a fundamental flaw present in the static minimax approach to understanding GANs.
As an alternative we explain GAN performance with ICR, a mechanism that focuses on the \emph{dynamics} of simultaneous training.
While there is more work left to be done in order to characterize ICR, we provide a number of illustrative experiments on low-dimensional examples and real GANs that supports our conclusions.
We also use a game-theoretic interpretation of ICR to identify algorithms such as CGD that can lead to stronger ICR.
Indeed, comprehensive experiments on CIFAR10 show systematically improved inception scores and stability when training with CGD, adding further support to our findings.\\
An important direction for future work is the closer investigation of the generator. 
Recent work on variational autoencoders \citep{razavi2019generating} and GANs  \citep{karras2019style} suggests that the inductive biases of the generator play an important role, as well.
Understanding their interaction with ICR is an important direction of future work. 
We also hope to better understand the relationship of ICR with local solution concepts such as \emph{``proximal equilibria''} \citep{farnia2020gans} that emphasize slow improvement of the discriminator.

\newpage
\subsubsection*{Acknowledgments}
We would like to thank Houman Owhadi for helpful discussions.
A. Anandkumar is supported in part by Bren endowed chair, DARPA PAIHR00111890035, Raytheon, and Microsoft, Google and Adobe faculty fellowships.
F. Sch{\"a}fer gratefully acknowledges support by  the Air Force Office of Scientific Research under award number FA9550-18-1-0271 (Games for Computation and Learning) and the Ronald and Maxine Linde Institute of Economic and Management Sciences at Caltech.
H. Zheng is supported by Zhiyuan College, Shanghai Jiao Tong University.

\bibliography{icr}
\bibliographystyle{icml2020}

\appendix
\section{Experimental details}

\subsection{Euclidean distance on images}
\begin{figure*}[ht]
    \centering
    \includegraphics[width=\textwidth]{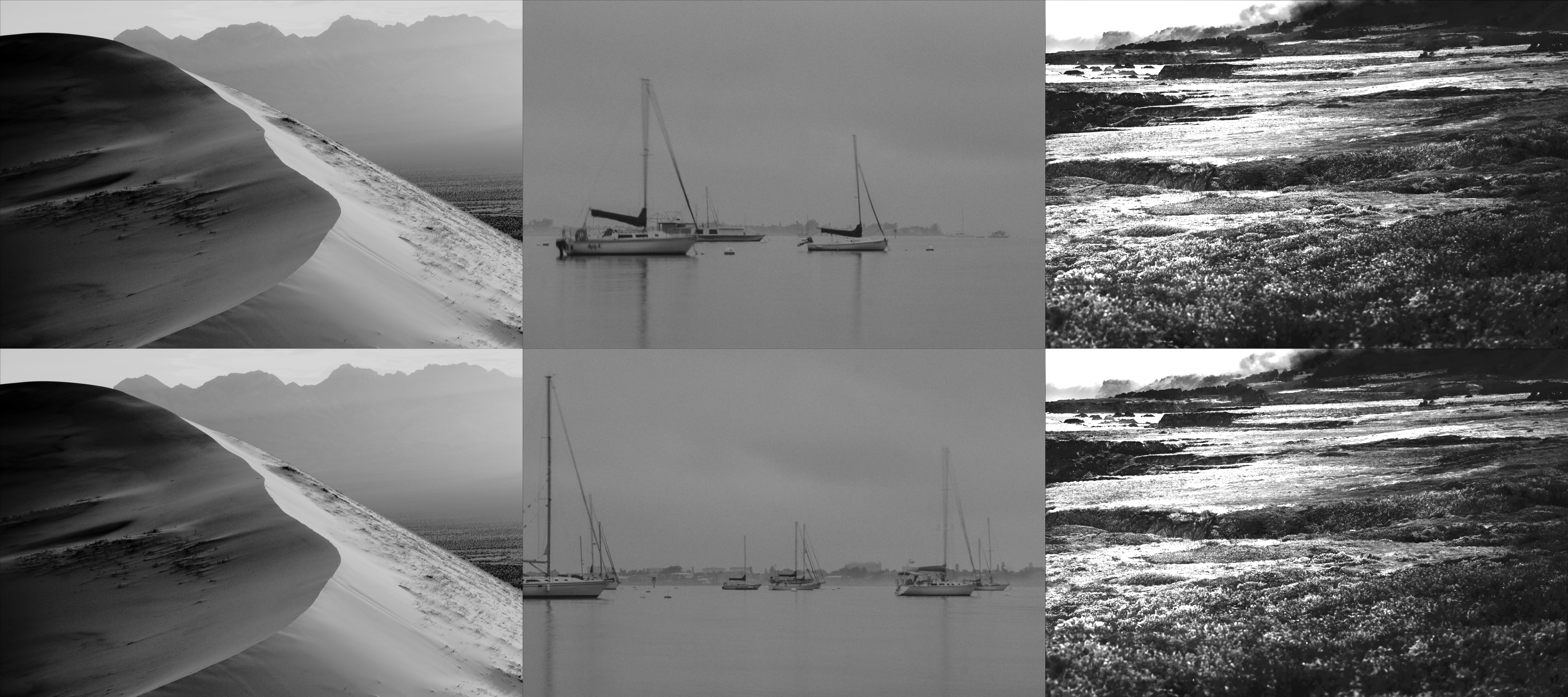}
    \caption{A larger reproduction of Figure~2 of the main paper. The first pair is based on an image by Matt Artz, the second pair on an image by Yanny Mishchuk, and the third pair on an image by Tim Mossholder. All images were obtained from \url{https://unsplash.com/}.}
    \label{fig:deception_reproduced}
\end{figure*}

In Figure~\ref{fig:deception_reproduced} we provide a larger reproduction of Figure~2 from the main paper. 
We see that also on the larger resolution, the third pair of images is visually indistinguishable, despite having the largest Euclidean distance of all pairs.
The textures of this image are very rough, with a rapid alternation of bright and dark pixels. 
Therefore, a slight warping of the image will lead to dark pixels taking the place of bright ones and vice versa, leading to a large Euclidean distance.
A similar effect could be achieved by the wind slightly moving the foliage between, for instance, successive frames of a video.
Thus, this phenomenon could be observed in real images.

\subsection{ICR as projection}
For the experiments in Figure 5 of the main paper, we used two tiny neural networks with 28 parameters and three layers each as generator and discriminator.

The generator $\mathcal{G}$ is composed as follows:
\begin{enumerate}[wide, labelwidth=!, labelindent=0pt, label=\textbf{\arabic*.}]
    \item Use first four parameters as input, apply arctan nonlinearity.
    \item Apply four times four dense layer, followed by arctan.
    \item Apply two times four dense layer, followed by the nonlinearity
    \begin{equation*}
        \begin{pmatrix}
        x \\
        y 
        \end{pmatrix} 
        \mapsto 
        \begin{pmatrix}
            \exp\left(\arctan\left(y\right) / \pi + x \right)\\
            \exp\left(\arctan\left(y\right) / \pi - x \right)
        \end{pmatrix}.
    \end{equation*}
\end{enumerate}

The form of the last nonlinearity ensures that the output is restricted to the set
\begin{equation*}
    \mathcal{S} \coloneqq \left\{ (e^{s+t}, e^{s-t}) \middle| s \in \left[-\frac{1}{2}, \frac{1}{2}\right], t \in \Reals \right\} \subset \Reals^2
\end{equation*}
that does not include the target $\Pd \coloneqq \left(2,2\right)$.
Note that in this simple example the generator does not take any input, but directly maps the weights $w_{\mathcal{G}} \in \Reals^{28}$ to a pair of real numbers.

The discriminator $\mathcal{D}_{\eta}$ is composed as follows:
\begin{enumerate}[wide, labelwidth=!, labelindent=0pt, label=\textbf{\arabic*.}]
    \item Rescale input by the diagonal matrix $\eta$, apply four times two dense layer, followed by arctan.
    \item Apply four times four dense layer, followed by arctan.
    \item Apply one times four dense layer, followed by arctan.
\end{enumerate}
While we did not observe the metastable projection behavior on all runs, we observed it in 13 out of 20 independent runs when using SimGD.
When using CGD we observed the projection behavior in 17 out of 20 independent runs (with the same initialization as in the SimGD cases).
Furthermore, the number of iterations spent in the projection states was larger when using CGD.

\subsection{ICR on MNIST}
In our experiments on MNIST, we use the network architectures detailed in Table \ref{table:mnist-g} and Table \ref{table:mnist-d}. We train using stochastic SimGD with a learning rate of 0.01. First,  we train the GAN for 9,000 iterations with a batch size of 128.  We refer to the resulting generator and discriminator as the checkpoint generator and discriminator. 
\subsubsection{Details about Figure 4 of the main paper}
\begin{figure}[htbp]
		\centering
		\subfigure[A part of the fake set]{
			\includegraphics[width=0.45\columnwidth]{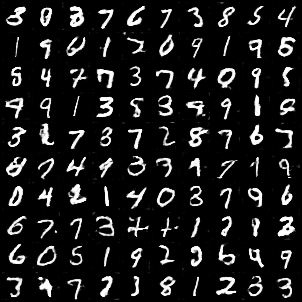}
		}
		\hfill
		\subfigure[A part of the real set]{
			\includegraphics[width=0.45\columnwidth]{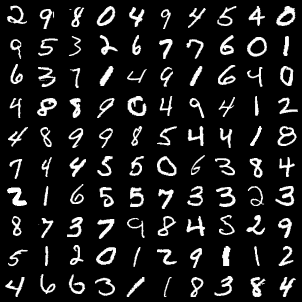}}
		\caption{Test set for Figure 4 of the main paper. }
		\label{fig:test set}
	\end{figure}

 We then create a test set that has a real set $X_{\mathrm{real}}$ that has 500 images sampled from MNIST training set, and a fake set $X_{\mathrm{fake}}$ that has 500 images generated by the checkpoint generator, as illustrated in Figure \ref{fig:test set}. 

Let $D_t, D_c$ denote, respectively, the discriminator at time step $t$ and the checkpoint discriminator.
The Euclidean distance between predictions of $D_t$ and $D_c$ over set $X_\mathrm{real}$ and $X_\mathrm{fake}$ in Figure 4 of the main paper is given by
\begin{equation}
\boldsymbol{d}_{\mathrm{set}}(D_t, D_c) = \sqrt{\sum_{x\in X_\mathrm{set}}(D_t(x)-D_c(x))^2}
\end{equation}
where $\mathrm{set} \in \{\mathrm{real}, \mathrm{fake}\}$.

\begin{table}[htbp]
	\begin{tabular}{c|c|c|c}
		\hline
		Module            & Kernel & Stride & Output shape \\
		\hline
		Gaussian distribution    & N/A    & N/A    & 96          \\
		Linear, BN, ReLU  & N/A    & N/A    & $1024$  \\
		Linear, BN, ReLU & N/A & N/A      & $128 \times 7 \times 7$  \\
		ConvT2d, BN, ReLU & $4\times 4$  & 2      & $64 \times 14 \times 14$ \\
		ConvT2d, Tanh     & $4\times 4$  & 2      & $1 \times 28 \times 28$  \\ \hline
	\end{tabular}
	\caption{Generator architecture for MNIST experiments}
	\label{table:mnist-g}
\end{table}

\begin{table}[htbp]
	\begin{tabular}{c|c|c|c}
		\hline
		Module            & Kernel & Stride & Output shape \\
		\hline
		Input  & N/A    & N/A    & $1 \times 28 \times 28$  \\
		Conv2d, LeakyReLU & $5\times 5$  & 1      & $32 \times 24 \times 24$  \\
		MaxPool & $2\times 2$  & N/A      & $32 \times 12 \times 12$ \\
		Conv2d, LeakyReLu & $5\times 5$  & 1      & $64\times 8 \times 8$  \\ 
		MaxPool & $2\times 2$  & N/A      & $64 \times 4 \times 4$ \\
		Linear, LeakyReLU & N/A    & N/A    & $1024$  \\
		Linear & N/A    & N/A    & $1$  \\
		\hline
	\end{tabular}
	\caption{Discriminator architecture for MNIST experiments}
	\label{table:mnist-d}
\end{table}

\subsection{CIFAR10 experiments}
\subsubsection{Architecture}
For our experiments on CIFAR10, we use the same DCGAN network architecture as in Wasserstein GAN with gradient penalty \citep{gulrajani2017improved}, which is reported in Table \ref{table:discriminator} and Table \ref{table:generator}. 

\subsubsection{Hyperparameters}
We compare the stability and performance of Adam and ACGD by varying the loss functions and regularization methods.

\textbf{Loss}: 
\begin{enumerate}[wide, labelwidth=!, labelindent=0pt, label=\textbf{\arabic*.}]
    \item  Original GAN loss \citep{goodfellow2014generative} $$ \mathcal{L}_o=\Expect_{x \sim \Pd} \left[ \log \mathcal{D}(x) \right] + \Expect_{x \sim \Pg}\left[ \log\left(1 - \D(x) \right) \right].$$
    \item Wasserstein GAN loss \citep{arjovsky2017wasserstein}
      $$\mathcal{L}_w=\Expect_{x \sim \Pd} \left[\mathcal{D}(x)\right] - \Expect_{x \sim \Pg} \left[\mathcal{D}(x)\right].$$
\end{enumerate}

\textbf{Regularization}: 
\begin{enumerate}[wide, labelwidth=!, labelindent=0pt, label=\textbf{\arabic*.}]
\item $L_2$ weight penalty on the discriminator parameters $\lambda \in \{10^{-2}, 10^{-3}, 10^{-4}\}$. \item Gradient penalty on the discriminator proposed by WGAN-GP paper \citep{gulrajani2017improved}. 
\item Spectral normalization on the discriminator proposed by SNGAN paper \citep{miyato2018spectral}. 
\end{enumerate}
Each experiment is trained with a batch size of 64. When using Adam and the original GAN loss, we adopt the log-trick as recommended in GAN paper \citep{goodfellow2014generative}. When using ACGD, the generator and discriminator share the same loss function. 
For the training of WGAN-GP, we use the same training strategy and hyperparameters as WGAN-GP\footnote{Check more details in WGAN-GP official repository: https://github.com/igul222/improved\_wgan\_training/blob/master/gan\_cifar.py} \citep{gulrajani2017improved}. Hyperparameter setting for each experiment is reported in Table \ref{table:detail}.

\begin{table*}[ht]
	\begin{tabular}{c|c|c|c|c|c|c|c}
		\hline
		Experiment        & Loss & Optimizer & Learning rate & Spectral Normalization & $L_2$ penalty & GP weight & Critic iterations \\
		\hline
		OGAN-0.0001L2+Adam & $L_o$ & Adam      & $10^{-4}$          & N/A     & $10^{-4}$       & N/A                     & 1                \\
		 
		OGAN-0.0001L2+ACGD & $L_o$ & ACGD      & $10^{-4}$          & N/A     & $10^{-4}$       & N/A                     & 1                 \\
		\hline
		OGAN-NOREG+Adam    & $L_o$ & Adam      & $10^{-4}$          & N/A     & N/A        & N/A                     & 1                 \\
	
		OGAN-NOREG+ACGD    & $L_o$ & ACGD      & $10^{-4}$          & N/A     & N/A        & N/A                     & 1                 \\
		\hline
		WGAN-GP+Adam       & $L_w$ & Adam      & $10^{-4}$          & N/A     & N/A        & 10                      & 5                 \\
		WGAN-SN+Adam       & $L_w$ & Adam      & $10^{-4}$          & Yes     & N/A        & N/A                      & 1                 \\
		WGAN-NOREG+ACGD    & $L_w$ & ACGD      & $10^{-4}$          & N/A     & N/A        & N/A                     & 1                 \\
		\hline
		WGAN-GP+ACGD       & $L_w$ & ACGD      & $10^{-4}$          & N/A     & N/A        & 10                      & 5                 \\
		WGAN-0.01L2+Adam   & $L_w$ & Adam      & $10^{-4}$          & N/A     & $10^{-2}$       & N/A                     & 1                 \\
		\hline
		WGAN-0.001L2+Adam  & $L_w$ & Adam      & $10^{-4}$          & N/A     & $10^{-3}$       & N/A                     & 1                 \\
		WGAN-0.001L2+ACGD  & $L_w$ & ACGD      & $10^{-4}$          & N/A     & $10^{-3}$       & N/A                     & 1                 \\
		\hline
	\end{tabular}
\caption{Settings for all the experiments that occurred in Figure 7 of the main paper. }
\label{table:detail}
\end{table*}

\subsubsection{Tensorflow inception score}
\begin{figure}
			\centering
			\subfigure[Against gradient call]{\includegraphics[width=0.49\columnwidth]{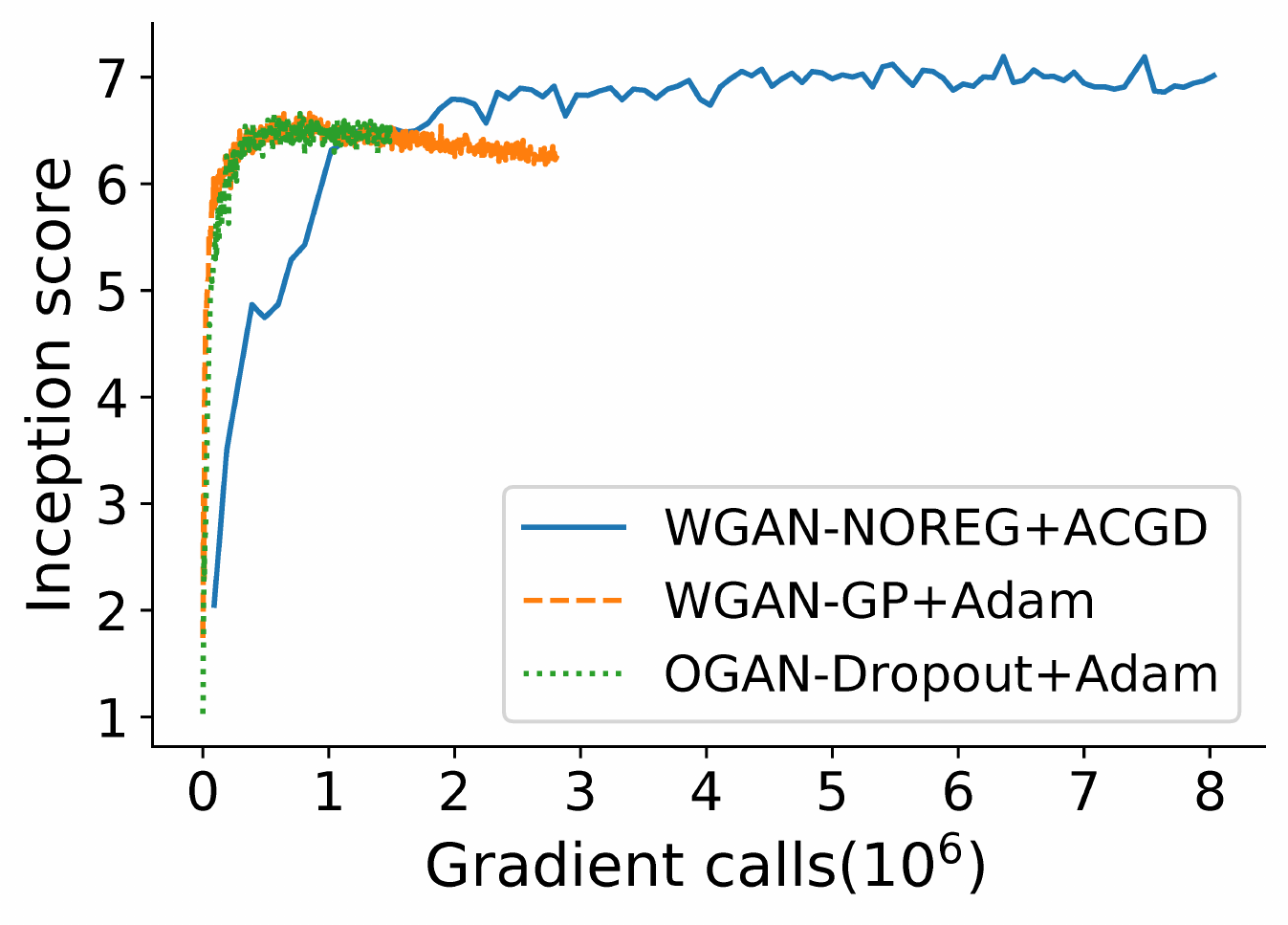}}
			\hfill 
			\subfigure[Against the generator iterations]{\includegraphics[width=0.49\columnwidth]{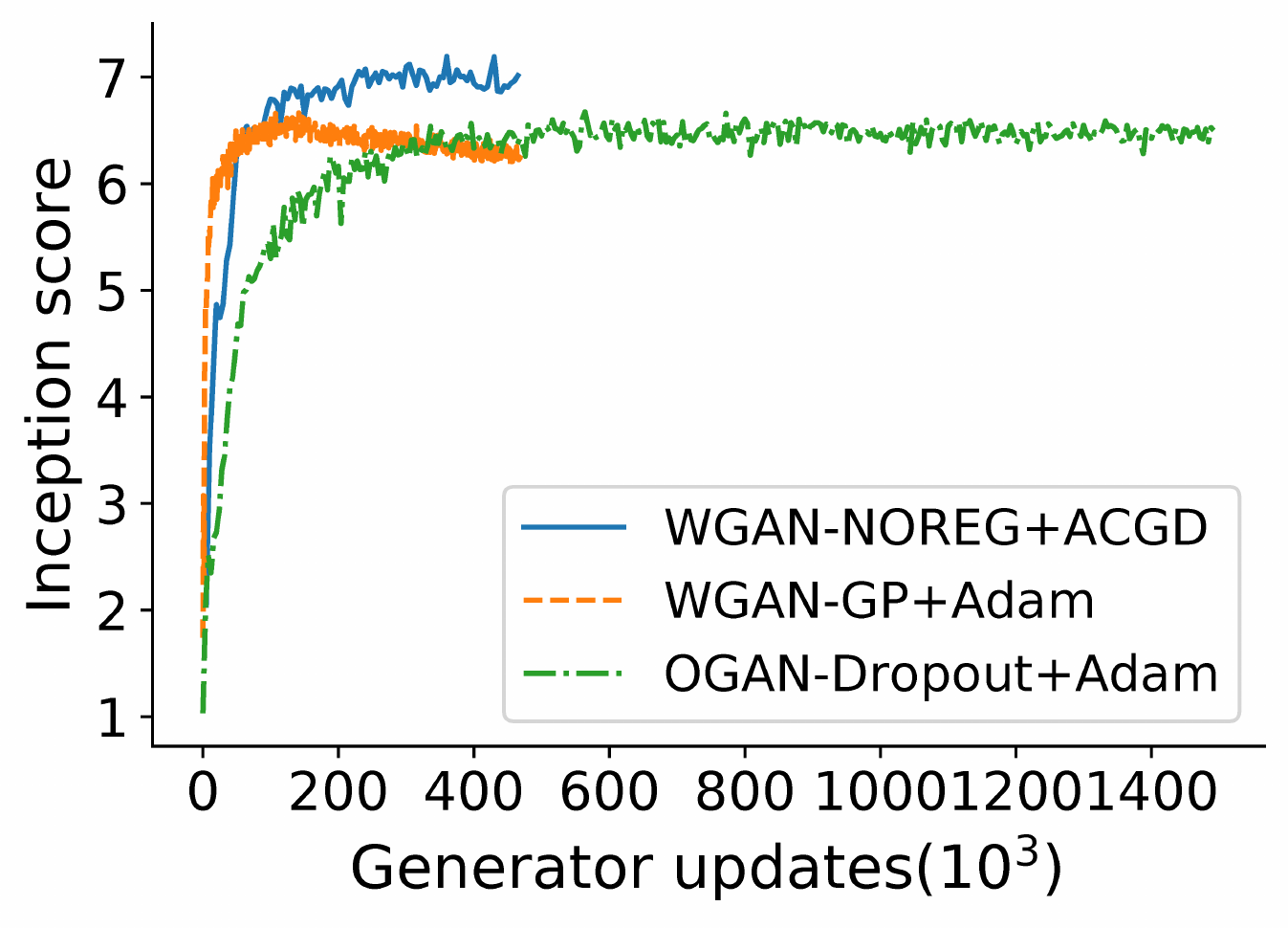}}
			
			\caption{Tensorflow inception scores for important runs}
			\label{fig:is-grad}
\end{figure}

We compute the Tensorflow version of the inception scores for important runs of our experiments to show that the relative performance of the different models is largely the same. As reported in Figure \ref{fig:is-grad}, our results matches the ones reported in the literature (Figure 3 in \citep{gulrajani2017improved}) with ACGD still outperforming WGAN-GP trained with Adam by around 10\%. 
 
\begin{table}[htbp]
	\begin{tabular}{c|c|c|c}
		\hline
		Module            & Kernel & Stride & Output shape \\
		\hline
		Gaussian distribution    & N/A    & N/A    & 128          \\
		Linear, BN, ReLU  & N/A    & N/A    & $256 \times 4 \times 4$  \\
		ConvT2d, BN, ReLU & $4\times 4$  & 2      & $128 \times 8 \times 8$  \\
		ConvT2d, BN, ReLU & $4\times 4$  & 2      & $64 \times 16 \times 16$ \\
		ConvT2d, Tanh     & $4\times 4$  & 2      & $3 \times 32 \times 32$  \\ \hline
	\end{tabular}
	\caption{Generator architecture for CIFAR10 experiments}
	\label{table:generator}
\end{table}

\begin{table}[htbp]
	\begin{tabular}{c|c|c|c}
		\hline
		Module            & Kernel & Stride & Output shape \\
		\hline
		
		Input  & N/A    & N/A    & $3 \times 32 \times 32$  \\
		Conv2d, LeakyReLU & $4\times 4$  & 2      & $64 \times 16 \times 16$  \\
		Conv2d, LeakyReLU & $4\times 4$  & 2      & $128 \times 8 \times 8$ \\
		Conv2d, LeakyReLu & $4\times 4$  & 2      & $256 \times 4 \times 4$  \\ 
		Linear & N/A    & N/A    & $1$  \\
		\hline
	\end{tabular}
	\caption{Discriminator architecture for CIFAR10 experiments}
	\label{table:discriminator}
\end{table}
\subsection{Details on ACGD}
In order to make a fair comparison with Adam, we run our experiments with ACGD, a variant of CGD that adaptively adjusts CGD's step size. The algorithm is described in Algorithm \ref{alg:acgd}. ACGD computes individual step sizes for the different parameters. Let $A_{x,t}$ and $A_{y,t}$ denote the diagonal matrices containing the step sizes of $x$ and $y$ at time step $t$ as elements. If $A_{x,t}$ and $A_{y,t}$ are multiples of the identity, the algorithm reduces to CGD with the corresponding step size. 
The reason we rearrange the terms as shown in Algorithm \ref{alg:acgd} is that we want the matrix inverse to contain an additive identity (to decrease the condition number) and be symmetric positive definite (so that we can use conjugate gradient \citep{eisenstat1981efficient} for its computation). 
ACGD adjusts CGD's step size adaptively during training with second moment estimate of the gradients. The update rules are derived from the local game in the same way as for CGD \citep{schaefer2019competitive}: 
\begin{align*}
&\min \limits_x x^{\top}\nabla_x f(x_t, y_t) + x^{\top} [D_{xy}f(x_t, y_t))] y +   \frac{1}{2}x^TA_{x,k}^{-1}x, \\
&\max \limits_y y^{\top} \nabla_y f(x_t, y_t) + y^{\top} [D_{yx}f(x_t, y_t))] x - \frac{1}{2}y^TA_{y,k}^{-1}y.
\end{align*}
	\begin{algorithm}[]
		\caption{ACGD, a variant of CGD with RMSProp-type heuristic to adjust learning rates. All operations on vectors are element wise. $D_{xy}^2f$, $D_{yx}^2f$ denote the mixed Hessian matrix $\frac{\partial^2f}{\partial x\partial y}$ and $\frac{\partial^2f}{\partial y \partial x}$. $\beta_2^t$ denotes $\beta_2$ to the power $t$. $\phi(\eta)$ denotes a diagonal matrix with $\eta$ on the diagonal. Hyperparameter settings for the tested GANs training problems are $\alpha =10^{-4}, \beta_2=0.99$, and $\epsilon = 10^{-5}$}
		\label{alg:acgd}
		\begin{algorithmic}
			\REQUIRE $\alpha$: Step size
			\REQUIRE $\beta_2$: Exponential decay rates for the second moment estimates
			\REQUIRE $\max_y \min_x f(x,y)$: zero-sum game objective function with parameters $x,y$
			\REQUIRE $x_0, y_0$ Initial parameter vectors
			
			$t \leftarrow 0$ Initialize timestep
			
			$v_{x,0},v_{y,0} \leftarrow 0$ (Initialize the $2^{nd}$ moment estimate)
			\REPEAT
			\STATE $t\leftarrow t+1$
			\STATE $v_{x,t}\leftarrow \beta_2\cdot v_{x,t-1}+(1-\beta_2) \cdot g_{x,t}^2$
			\STATE $v_{y,t} \leftarrow \beta_2 \cdot v_{y,t-1}+(1-\beta_2)\cdot g_{y,t}^2$
			
			\STATE $v_{x,t} \leftarrow v_{x,t} / (1-\beta_2^t)$
			\STATE $v_{y,t} \leftarrow v_{y,t} / (1-\beta_2^t)$ (Initialization bias correction )
			
			\STATE $\eta_{x,t} \leftarrow \alpha /(\sqrt{v_{x,t}}+\epsilon)$
			\STATE $\eta_{y,t} \leftarrow \alpha/(\sqrt{v_{y,t}}+\epsilon)$
			
			\STATE $A_{x,t} = \phi(\eta_{x,t})$
			\STATE $A_{y,t} = \phi(\eta_{y,t})$
			
			\STATE $\Delta x_t \leftarrow - A_{x,t}^{\frac{1}{2}}(I+A_{x,t}^{\frac{1}{2}}D_{xy}^2f A_{y,t}D_{yx}^2f A_{x,t}^{\frac{1}{2}})^{-1}A_{x,t}^{\frac{1}{2}}$\\$
			(\nabla_x f + D_{xy}^2fA_{y,t}\nabla_yf)$
			\STATE $\Delta y_t \leftarrow  A_{y,t}^{\frac{1}{2}}(I+A_{y,t}^{\frac{1}{2}}D_{yx}^2f A_{x,t}D_{xy}^2f A_{y,t}^{\frac{1}{2}})^{-1}A_{y,t}^{\frac{1}{2}}$\\$
			(\nabla_y f - D_{yx}^2fA_{x,t}\nabla_xf)$
			\STATE $x_t\leftarrow x_{t-1} + \Delta x_t $
			\STATE $y_t\leftarrow y_{t-1} + \Delta y_t$

			\UNTIL{$x_t,y_t$ converged}
		\end{algorithmic}
	\end{algorithm}

\end{document}